\documentclass[10pt,journal,twoside]{IEEEtran}
\IEEEoverridecommandlockouts 

\usepackage[bookmarks=false,hidelinks]{hyperref}

\newcommand{\ph}[1]{#1}

\usepackage{balance}
\usepackage{cite}
\usepackage{amsmath,amssymb,amsfonts}
\usepackage[utf8]{inputenc}
\usepackage{textcomp}
\usepackage{xcolor}
\usepackage{booktabs}
\usepackage{makecell}
\usepackage{algpseudocode}
\usepackage{multirow}
\usepackage{algorithm}
\usepackage{soul}
\usepackage{graphicx}
\usepackage{hyperref}
\usepackage{caption}
\usepackage{subfigure} 

\def\BibTeX{{\rm B\kern-.05em{\sc i\kern-.025em b}\kern-.08em
    T\kern-.1667em\lower.7ex\hbox{E}\kern-.125emX}}

\begin{document}

\markboth{Preprint}%
{Nguyen et al.: TinySense+: Entropy-Aware CSI Compression for Scalable and Adaptive Wi-Fi Sensing}

\title{$\mathtt{TinySense+}$: Entropy-Aware CSI Compression for Scalable and Adaptive Wi-Fi Sensing}

\author{Phuc H. Nguyen, Toan Gian, Cuong D. Do, Symeon Chatzinotas and Van-Dinh Nguyen\\ \thanks{P. H. Nguyen T. Gian and C. D. Do are with the Smart Green Transformation Center (GREEN-X), and College of Engineering \& Computer Science, VinUniversity, Hanoi, Vietnam (e-mail: \{23phuc.nh, toan.gd, cuong.dd\}@vinuni.edu.vn). S. Chatzinotas is with SnT, University of Luxembourg, Luxembourg (e-mail: symeon.chatzinotas@uni.lu). V.-D. Nguyen is with the School of Computer Science and Statistics, Trinity College Dublin, Ireland (e-mail: dinh.nguyen@tcd.ie). An earlier version of this paper was presented in part at \textit{IEEE Int. Conf. Perva. Comput. and Commun. (PerCom)}, March 2026 \cite{TinySensePerCom2026}. \vspace{-15pt}} 

}
\maketitle

\begin{abstract}
Wi-Fi human pose estimation requires high-rate channel state information (CSI), making communication a bottleneck when sensing is offloaded from resource-constrained devices. Effective compression must therefore preserve pose information while accommodating different communication budgets. We present $\mathtt{TinySense+}$, an efficient compression framework for scalable Wi-Fi-based human sensing with entropy-aware transmission and test-time rate adjustment. The framework uses a vector-quantized generative adversarial network to encode CSI into discrete latent indices shared between an edge device and a server. A causal Transformer predicts conditional index probabilities for arithmetic coding, while a differentiable rate--distortion objective balances expected code length, CSI reconstruction, and pose accuracy. Prefix-conditioned latent completion reuses the entropy model to infer an omitted suffix, allowing the transmitted prefix length to change without retraining the encoder. We evaluate the framework on MM-Fi and WiPose datasets and prototype it on Raspberry Pi and Jetson Nano devices connected to an RTX 8000 server. On MM-Fi, entropy coding reduces the baseline cost from $8.00$ to $4.82$ bits per index without changing pose accuracy. Rate-aware training further reduces this cost to $3.21$ bits per index and improves PCK$_{20}$ from $41.21\%$ to $44.01\%$. Results also show that transmitting half of the latent sequence retains a PCK$_{20}$ of $40.36\%$. The evaluated operating range reaches a compression ratio of $5840$ with PCK$_{20}=37.92\%$. These results show that optimizing latent predictability alongside sensing fidelity provides a useful communication--accuracy trade-off for edge--server Wi-Fi sensing. The code repository is available online at \href{https://github.com/phucngvinuni/TinySensePlus}{GitHub}.
\end{abstract}

\section{Introduction}
\subsection{Motivation}\label{sub_IA}
Wi-Fi sensing uses changes in channel state information (CSI) to infer human activities without requiring people to wear sensors or appear in camera images. The use of commodity wireless infrastructure has enabled device-free localization, activity recognition, gesture recognition, and human pose estimation (HPE)~\cite{9,10,11, Wang2019CanWE, Wipose, csipose,realtime}. For continuous HPE, however, sensing accuracy is only one requirement: the system must also operate within the communication, computation, and energy budgets of the sensing device.

Deep models such as MetaFi++~\cite{MetaFi++} improve pose estimation but can be costly to run locally. Offloading computation to a server reduces this burden while introducing a different bottleneck: transmitting high-dimensional CSI at a high sampling rate. For example, $114$ subcarriers, three antenna links, and $500$ samples/s require $10.944$~Mbit/s when each complex coefficient is represented by two $32$-bit values. Even amplitude-only transmission requires $5.472$~Mbit/s under these assumptions. This traffic grows with the number of sensing links, making CSI compression essential for reducing the communication cost of offloading pose estimation to a server.

CSI compression must preserve information needed for pose estimation. The low reconstruction error alone does not ensure that small motion-dependent variations remain useful to the downstream estimator. Existing sensing-oriented compression systems address parts of this problem~\cite{19,20,AutoCompress}, but the transmitted representation must also support different communication budgets. Our conference paper, $\mathtt{TinySense}$~\cite{TinySensePerCom2026}, addressed this need through task-aware vector quantization (VQ), K-means codebook resizing, and missing-index recovery. The fixed-length index representation nevertheless assigns the same number of bits to common and rare symbols, leaving statistical redundancy unused. Codebook resizing also provides only a discrete set of operating configurations.

This problem is particularly relevant to continuous edge--server sensing. Communication repeats for every sensing window, whereas the device must also acquire samples and run the encoder. Transmitting less data can reduce communication cost, but an overly aggressive representation may remove the weak channel variations needed to distinguish joint locations. Moving all processing to the edge avoids offloading but places the pose estimator on a resource-constrained device. A useful compression system must therefore preserve task information and expose operating points that balance payload, inference cost, and pose accuracy. Reconstruction fidelity alone cannot establish this balance.

$\mathtt{TinySense}$ leaves two connected gaps. First, fixed-length indices do not reward an encoder for producing predictable symbol sequences: different representations with the same codebook size and index-sequence length incur the same nominal cost. Second, changing the codebook changes the representation itself, whereas a deployed system may need to reduce its next transmission while retaining the current encoder and codebook. $\mathtt{TinySense+}$ addresses these gaps through entropy-aware training and prefix-conditioned completion, linking representation learning to coded length and enabling shortened transmissions. The representation is learned from CSI–pose pairs for HPE and may depend on the training environments; generalization to unseen sites without adaptation remains unverified.

These limitations motivate two questions: \textit{can the latent representation be optimized for both sensing accuracy and its coded length, and can a deployed model change how much information it sends without retraining?} The learned entropy models and rate-aware vector quantization offer useful tools~\cite{ball2018variational,NEURIPS2018_53edebc5,qian2022entroformer,rdvq}. Their value for Wi-Fi HPE depends on the resulting pose accuracy and on whether communication savings justify the added computation at the edge and server.

\begin{figure}[t]
  \centering
  \includegraphics[width=0.95\linewidth]{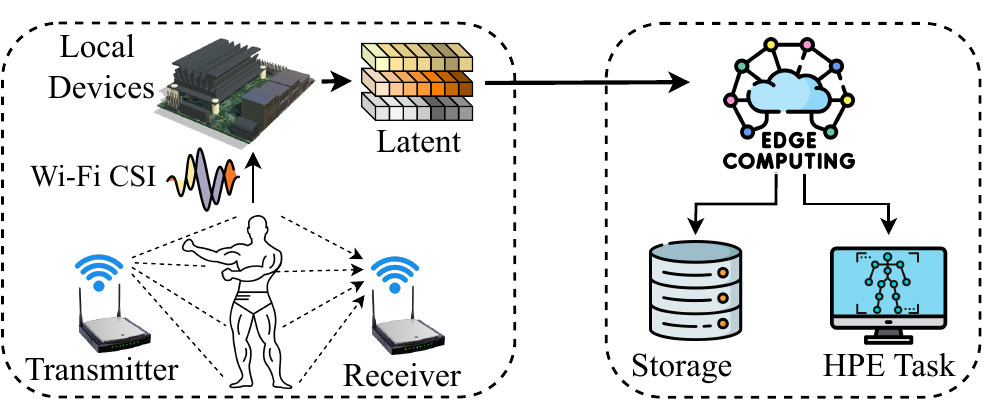}
  \caption{Edge--server Wi-Fi sensing: local devices compress
CSI into latent indices for transmission and server-side pose estimation.}
  \label{fig:figure1}
\end{figure}

\subsection{Main Contributions}
We present $\mathtt{TinySense+}$, an entropy-aware extension of $\mathtt{TinySense}$ for edge--server Wi-Fi sensing, as illustrated in Fig.~\ref{fig:figure1}. $\mathtt{TinySense}$ ~\cite{TinySensePerCom2026} established the vector-quantized generative adversarial network (VQGAN) sensing representation, fixed-length transmission, K-means rate selection, and index-recovery module. This work extends that system with learned entropy coding, a differentiable rate objective, and prefix-conditioned latent completion (PCLC), together with component ablations, rate-sensitivity experiments, prefix sweeps, and additional deployment measurements. The extension connects task-aware representation learning to actual coded payload and test-time transmission length.

\begin{table}[H]
\centering
\caption{TinySense and TinySense+: Key Differences}
\label{tab:extension_summary}
\small
\begin{tabular}{@{}
>{\raggedright\arraybackslash}p{0.22\columnwidth}
>{\raggedright\arraybackslash}p{0.30\columnwidth}
>{\raggedright\arraybackslash}p{\dimexpr0.42\columnwidth-12pt\relax}
@{}}
\toprule
Aspect & $\mathtt{TinySense}$ & $\mathtt{TinySense}+$ \\
\midrule
Training & Task-aware VQGAN & Adds rate regularization \\ 

Index coding & Fixed-length binary & Causal entropy model with arithmetic coding \\ 

Rate control & Codebook resizing & Adds adjustable prefix length \\ 

Recovery & Predicts erased indices & Entropy model to complete omitted suffix \\ 

Evaluation & Compression, sensing, and testbed costs & Adds rate and prefix ablations \\
\bottomrule
\end{tabular}
\end{table}

Table~\ref{tab:extension_summary} summarizes the relationship between the two approaches. The new significance is the adaptation and evaluation of these mechanisms for CSI-based HPE under edge--server constraints. In particular, the differentiable VQ rate optimization and prefix completion build on ideas studied for image compression in rate--distortion-optimized vector quantization (RDVQ)~\cite{rdvq}, where they are combined with pose supervision and evaluated through sensing and system metrics. 
The significance of this integration is that it separates three controls that were coupled or absent in $\mathtt{TinySense}$ (our conference version) representational capacity through the codebook, statistical coding cost through the entropy model, and transmitted context through the prefix length. Their interaction is not automatically beneficial for sensing. A representation that is easier to predict can discard useful motion information, and a shorter prefix can increase completion error and server computation. $\mathtt{TinySense}+$ therefore tests these mechanisms through pose metrics and component ablations, and examines whether payload savings translate into practical benefits on the evaluated edge platforms. This work extends the sensing system and evaluates its trade-offs between communication cost and sensing accuracy.

The main contributions are summarized as follows.
\begin{itemize}
\item We develop $\mathtt{TinySense}$ to compress CSI into discrete latent indices using a shared VQGAN codebook while preserving CSI reconstruction and HPE accuracy. A lightweight encoder runs on the edge device, while reconstruction and pose estimation run on the server, reducing local computation and transmitted data.

\item We use K-means clustering to derive smaller codebooks from a pretrained codebook, providing different bitrate--accuracy operating points. A recovery Transformer predicts unavailable indices from visible context, improving sensing performance under the evaluated index-erasure conditions.

\item We extend the base framework to $\mathtt{TinySense+}$ by combining a causal entropy Transformer with arithmetic coding. A differentiable rate term complements reconstruction and keypoint supervision, encouraging latent indices that preserve sensing information and require fewer bits to transmit.

\item We transmit a selected prefix of the latent indices and reuse the causal entropy model to complete the omitted suffix at the server. The prefix length controls the communication--accuracy trade-off without encoder retraining or codebook resizing.

\item We evaluate both versions on MM-Fi and WiPose and deploy them on Raspberry Pi and Jetson Nano devices with an RTX 8000 server. At $\eta=1710$ on MM-Fi, $\mathtt{TinySense}$ achieves a percentage of correct keypoints at threshold $20$ (PCK$_{20}$) of $32.93\%$, versus $25.62\%$ for EfficientFi. $\mathtt{TinySense+}$ reaches $\eta=5840$ with PCK$_{20}=37.92\%$. Component ablations and testbed measurements characterize coding, pose-accuracy, latency, energy, and payload trade-offs.
\end{itemize}

The remainder of this paper is organized as follows. Section~\ref{sec:related-work} reviews related work, and Section~\ref{sec:background} provides the necessary background. Sections~\ref{sec:methodology} and~\ref{sec:TinySense+} describe the core system design and its enhancements, respectively. Section~\ref{sec:experimental-result} presents the experimental evaluation, and Section~\ref{sec:conclusion} concludes the paper.

\section{Related Work}\label{sec:related-work}
\textbf{Wi-Fi-based pose estimation.} Early systems inferred human poses from commodity Wi-Fi measurements using learned spatial and temporal features~\cite{Toward2020,Guo2020,Wang2019CanWE}. MetaFi and MetaFi++~\cite{MetaFi,MetaFi++} developed this direction for avatar applications, while PerUnet~\cite{Wipose} introduced channel attention. AdaPose~\cite{AdaPose} addressed cross-site variation, PiW-3D~\cite{PiW3D} considered multiple people, and CSIPose~\cite{csipose} studied through-wall estimation. Lightweight architectures~\cite{me,toan,toan1,toan2} reduced inference costs, complementing broader sensing benchmarks and edge-system analyses~\cite{13,14}. These advances primarily concern estimation or local processing. They do not by themselves determine how many bits an edge device must transmit. $\mathtt{TinySense+}$ addresses this complementary requirement by optimizing the representation used for both communication and HPE.

\textbf{CSI compression for communication and sensing.} Linear dimensionality reduction and sparse recovery~\cite{FRS1901LIIIOL,Eckart1936TheAO,29} provide compact signal representations, while learned feedback methods such as CsiNet~\cite{30} and DeepCMC~\cite{31} exploit channel structure. Sensing-oriented systems move beyond channel reconstruction: EfficientFi~\cite{19} uses discrete representations, RSCNet~\cite{20} supports dynamic compression, and AutoCompress~\cite{AutoCompress} selects informative spectrogram elements under bandwidth constraints. Our conference paper, $\mathtt{TinySense}$~\cite{TinySensePerCom2026}, combined pose-aware VQGAN training, codebook resizing, and index recovery. However, its fixed-length coding leaves index probabilities unused, and a small reconstruction error alone does not guarantee accurate pose estimates. $\mathtt{TinySense+}$ adds conditional entropy coding and rate regularization to a pose-supervised representation, explicitly linking coded length to sensing quality.

\textbf{Learned entropy coding and rate adaptation.} Learned image codecs combine latent probability models with entropy coding~\cite{ball2018variational,NEURIPS2018_53edebc5}. Entroformer~\cite{qian2022entroformer} used a Transformer to model contextual dependencies. Variable-rate autoencoders and code editing offer further control over compression~\cite{50,51,52}. More closely related, RDVQ~\cite{rdvq} coupled soft vector-quantization assignments with an autoregressive entropy model and supports prefix transmission followed by latent completion. These methods establish mechanisms for image compression, but their image-fidelity objectives do not directly measure pose information in CSI or deployment costs on sensing devices. We adapt these mechanisms to the $\mathtt{TinySense}$ pipeline with keypoint supervision and evaluate their communication--accuracy trade-off alongside edge--server latency and energy.

Taken together, these studies motivate a task-aware evaluation of compression rather than treating image fidelity, channel reconstruction, and pose accuracy as equivalent objectives. Existing learned codecs supply the entropy-modeling and rate-control mechanisms, while sensing systems supply the task supervision and deployment setting. The remaining question addressed here is how their combination changes transmitted cost, pose quality, and computation within one CSI-based pipeline. By distinguishing the inherited components in $\mathtt{TinySense}$ from the added coding, training, and prefix mechanisms, the evaluation can attribute the benefits more precisely and delimit the conditions under which they hold.

\section{Background} \label{sec:background}
\subsection{CSI and Its Limitations}\label{AA}
CSI captures fine-grained information about the wireless transmission channel, reflecting environmental characteristics and human motion through multipath propagation and signal reflections. Modern IEEE~802.11 Wi-Fi systems employ Orthogonal Frequency-Division Multiplexing (OFDM) and multiple antennas, enabling CSI to record amplitude attenuation and phase shifts across subcarriers and antenna pairs. CSI samples the channel frequency response at the OFDM subcarriers. The corresponding delay-domain multipath model is expressed by the following channel impulse response
\begin{equation}
h(\tau)=\sum\nolimits_{l=1}^{L}\alpha_l e^{j\phi_l}
\delta(\tau-\tau_l)
\end{equation}
where $\tau$ is the delay variable, and $\alpha_l$, $\phi_l$, and $\tau_l$ denote the amplitude, phase, and propagation delay of the $l$-th multipath component, respectively. $L$ is the number of multipath components, and $\delta(\cdot)$ denotes the Dirac delta function.

The communication cost depends on the number of subcarriers, antenna links, sampling rate, and numerical representation. Section~\ref{sub_IA} gives an example for complex-valued and amplitude-only CSI. The experimental compression ratios use the amplitude-only reference defined in Section~\ref{sec:experimental-result}.

\subsection{VQ-VAE in Data Compression}
A variational autoencoder (VAE)~\cite{kingma2013auto} models the data distribution through a latent variable $z$, with joint distribution $p(x,z)=p(x\mid z)p(z)$. Because computing the marginal likelihood is generally intractable, the encoder uses a variational posterior $q(z\mid x)$ to approximate $p(z\mid x)$. The encoder and decoder jointly maximize the evidence lower bound (ELBO), which equals the data log-likelihood when the approximate and true posteriors coincide:
\begin{align}
\log p(x) &\geq \text{ELBO}(x) \notag\\
&= \underbrace{\mathbb{E}_{q(z \mid x)} \big[ \log p(x \mid z) \big]}_{\text{Distortion}} - \underbrace{D_{\mathrm{KL}}\!\big(q(z \mid x)\,\|\,p(z)\big)}_{\text{Rate}}
\label{eq:elbo}
\end{align}
where the negative expected log-likelihood corresponds to reconstruction distortion, while the rate term regularizes the latent representation by encouraging the variational posterior $q(z \mid x)$ to remain close to the prior $p(z)$. A vector-quantized variational autoencoder (VQ-VAE)~\cite{55} is a variant of the VAE~\cite{rezende2014stochastic} that employs discrete latent variables, even for continuous inputs. Owing to the fixed size and discrete nature of its latent code, VQ-VAE naturally lends itself to lossy compression. Discrete latent representations are also used in music generation and high-resolution image synthesis.

\subsection{Entropy Coding} \label{sec:entropy_coding}
In information theory, the theoretical lower bound of lossless data compression is characterized by Shannon entropy~\cite{6773024}. For a discrete random variable $X$ with probability mass function $p(x)$ over a finite alphabet $\mathcal{X}$, its entropy $H(X)$ is defined as
\begin{equation}
H(X) = - \sum_{x \in \mathcal{X}} p(x) \log_2 p(x).
\end{equation}
While conventional quantization-based compression often represents each index using a fixed-length binary code, requiring $\lceil \log_2 K \rceil$ bits for a codebook of size $K$, this strategy can be highly suboptimal when the latent index distribution is skewed. Entropy coding exploits symbol probabilities to reduce bit cost without altering transmitted indices. Task relevance is learned through keypoint supervision, while rate regularization encourages compressibility. In particular, arithmetic coding (AC)~\cite{286189} encodes an entire symbol sequence as a single fractional value within the interval $[0,1)$. By recursively partitioning this interval in proportion to the symbol probabilities, AC can achieve an average coding rate close to the entropy limit. The effectiveness of AC, however, depends critically on the accuracy of the estimated probability distribution. In modern neural compression systems~\cite{ball2018variational}, deep autoregressive models~\cite{NEURIPS2018_53edebc5} are commonly used as entropy models to estimate context-aware conditional probabilities. Recent studies, such as Entroformer~\cite{qian2022entroformer}, further demonstrate that Transformer-based entropy models can effectively predict $p(x_i\mid x_{<i})$ for discrete latent sequences, enabling more efficient entropy coding.

\section{TinySense: Fixed-length Compression} 
\label{sec:methodology}

\begin{figure*}[tb]
  \centering
  \includegraphics[width=0.88\linewidth]{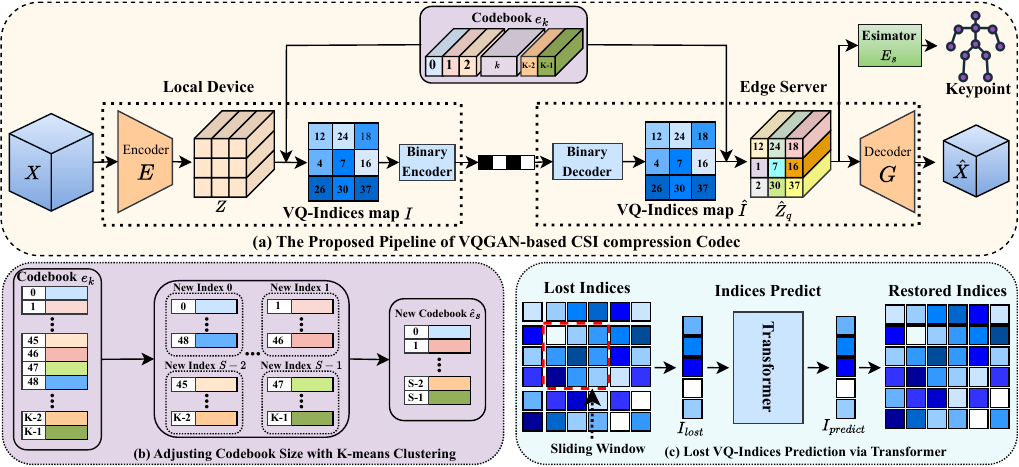}
  \caption{Illustration of the proposed $\mathtt{TinySense}$ framework: (a) the overall compression pipeline, (b) the K-means clustering algorithm
to adjust the codebook size for variable bitrates, and (c) a transformer approach for predicting lost VQ indices. \vspace{-0.3cm}}
\label{fig:figure2}
\end{figure*}

Our approach consists of $i$) an edge model at the local device and $ii$) a cloud model (see Fig.~\ref{fig:figure2} (a)). The edge encoder $E$ extracts latent features $\mathbf{Z}$ from CSI $\mathbf{X}$ under keypoint supervision. Compression is therefore task-aware, while fixed-length coding represents the resulting indices without using task labels. These features are then compressed to reduce data size using a specified compression ratio $\eta$, with $\textbf{Z}$ mapped to VQ-indices $\textbf{I}$ through nearest-neighbor matching in a CSI codebook $\textbf{e}_k$ (Sec. \ref{sec:VQ-compression}) and inverted back to a quantized vector $\textbf{Z}_q$. The indices $\textbf{I}$ are represented by fixed-length binary words and transmitted to the server. On the server, $\mathtt{TinySense}$ decompresses and reconstructs $\hat{\textbf{Z}}_q$ by referencing the shared codebook $\textbf{e}_k$ to retrieve the codewords identified by the received index map $\hat{\textbf{I}}$, enabling two primary tasks: $i)$ HPE via an Estimator $E_s$ and $ii)$ CSI reconstruction using Decoder $G$. For CSI data \textbf{X} $\in$ $\mathbb{R}^{F \times T  \times C}$, where $F$, $T$, and $C$ denote the numbers of frequency bins, time samples, and channels, respectively, the compression pipeline includes five key components:
\begin{itemize}
   \item \textbf{Encoder $E(\cdot; \theta_E)$} transforms the input signal \textbf{X} $\in$ $\mathbb{R}^{F \times T  \times C}$ into a latent representation $\textbf{Z}$ $\in$ $\mathbb{R}^{\frac{F}{M} \times \frac{T}{M}  \times D}$, where $M$ and $D$ are the encoder downsampling factor and the dimensionality of the latent codes, respectively.
  
  \item \textbf{Codebook} $\mathbf{e}_k \in \mathbb{R}^{D},\forall k \in \{1,2,\cdots, K$\} maps the latent representation $\textbf{Z}$ to a sequence of VQ indices $\textbf{I}$ and reconstructs it into a quantized version $\textbf{Z}_{q}$ $\in$ $\mathbb{R}^{\frac{F}{M} \times \frac{T}{M}  \times D}$ using the nearest neighbor lookup. A K-means clustering algorithm derives smaller codebooks from cluster centroids (see Fig. \ref{fig:figure2}(b)).

  \item \textbf{Decoder  $G(\cdot; \theta_G)$} acts as a generator within the generative adversarial network (GAN) framework, converting the quantized latent representation $\hat{\textbf{Z}}_q$ back into reconstructed CSI data $\hat{\textbf{X}}$ $\in$ $\mathbb{R}^{F \times T  \times C}$. 

  \item \textbf{Estimator $E_s(\cdot; \theta_R)$}  produces HPE based on the quantized latent representations $\hat{\textbf{Z}}_q$.

  \item \textbf{Transformer $T$} predicts missing indices using the context of existing indices, as shown in Fig.~\ref{fig:figure2}~(c).
  
\end{itemize} \vspace{-0cm}

\subsection{VQ-indices Compression}\label{sec:VQ-compression}
Unlike existing compression methods \cite{48,49,50,51,52} that adopt scalar quantization, we integrate convolutional neural network (CNN) inductive biases and principles from neural discrete representation learning \cite{26}. \smallskip

\noindent \textbf{VQ-indices process.}~Our model jointly learns an encoder $E$, a decoder $G$, and a codebook of $K$ trainable vectors $\mathbf{e}_k\in\mathbb{R}^{D}$, which is shared by the local device and server. During training, the codebook loss updates the selected vectors toward their assigned encoder outputs. Each latent vector $\mathbf{Z}_{ij}$ is then represented by the index of its nearest codeword:
\begin{equation}
    \mathbf{I}_{ij}=\underset{k\in1,2,...,K}{\mathtt{argmin}}\|\textbf{Z}_{ij}-\textbf{e}_{k}\| \label{eq:1}
\end{equation}
where $i$ and $j$ denote the positions of vector $\textbf{Z}_{ij}$, with $\textbf{I}_{ij}$ representing the corresponding index. We obtain $\textbf{Z}_q$ through the quantization function $\textbf{q}(\cdot)$, which replaces each latent vector $\textbf{Z}_{ij}$ with the codeword selected by $\textbf{I}_{ij}$
\begin{equation}
    \textbf{Z}_{q}=\textbf{q}(\textbf{Z})=\{\textbf{e}_{k}\mid\textbf{I}_{ij}\}\in\mathbb{R}^{\frac{F}{M} \times \frac{T}{M} \times D} \label{eq:1a}.
\end{equation}
\textcolor{black}{We then apply the Binary encoder to convert VQ indices ($i.e.,$ decimal values) \textbf{I} into bitstreams ($i.e.,$ binary values) for transmission to the nearest server}.\smallskip

\noindent \textbf{VQ loss function.}~After decoding, the reconstructed latent vectors $\hat{\textbf{Z}}_q$ are generated by retrieving their corresponding codewords based on their indices. The decoder $G$ synthesizes the reconstructed CSI data $\hat{\textbf{X}}\thickapprox \textbf{X}$ as
\begin{equation}
    \hat{\textbf{X}} = G(\textbf{Z}_q) = G\big(\textbf{q}(E(\textbf{X}))\big). \label{eq:2}
\end{equation}
Backpropagation through the non-differentiable quantization operation in \eqref{eq:2} is managed with a straight-through gradient estimator. This allows gradients to flow from the decoder to the encoder \cite{57} and enables end-to-end training of the model and codebook via the following loss function
\begin{align}
    \mathcal{L}_{\mathtt{VQ}}(E, G, \hat{e}_s) = & \, \|\mathbf{X} - \hat{\mathbf{X}}\|_2^2 + \|\mathtt{sg}[E(\textbf{X})] - \textbf{Z}_q\|^2 \nonumber\\
    &  + \beta\|\mathtt{sg}[\textbf{Z}_q] - E(\textbf{X})\|_2^2
     \label{eq:3}
\end{align}
where $\mathtt{sg}[\cdot]$ indicates the stop-gradient operation, and 
$\|\mathtt{sg}[\textbf{Z}_q] - E(\textbf{X})\|_2^2$ is the ``\textit{commitment loss}'' with weighting factor $\beta$ \cite{55}. The codebook term $\|\mathtt{sg}[E(\mathbf{X})]-\mathbf{Z}_q\|^2$ fits the selected codewords to the encoder features, while the commitment term keeps the encoder outputs close to their assigned codewords.\vspace{3pt}

\noindent\textbf{Adaptive compression via K-means.}~Codebook design is critical to compression efficiency. We implement a rate control strategy that adapts the compression bitrate by resizing the codebook using K-means clustering. We first learn a large codebook jointly with the encoder, decoder, and pose estimator during offline training. K-means then clusters the learned codewords, and the resulting centroids initialize smaller codebooks for fine-tuning.

K-means initializes centroids from the pretrained codebook $\mathbf{e}_k$ and assigns each codeword to its nearest centroid as
\begin{equation}
    \textbf{C}_{s}=\underset{s\in1,2,...,S}{\mathtt{argmin}}\|\textbf{e}_{k}-\hat{\textbf{e}}_{s}\|^2\label{eq:4}
\end{equation}
where $\textbf{C}_s$ represents the $s$-th cluster, $\hat{\textbf{e}}_s$ denotes the centroid of that cluster, and $S$ (with $S < K$) denotes the new codebook size. Following this, the mean of the codebook vectors within each cluster $C_j$ is recalculated, updating the cluster centroids accordingly as $\hat{\mathbf{e}}_s = \frac{1}{|C_s|} \sum_{\mathbf{e} \in C_s} \mathbf{e}$.
These steps are repeated iteratively until the clustering minimizes the associated cost function
\begin{equation}
    \min J(\mathbf{e}_k; \hat{\mathbf{e}}_s) = \min \Big( \frac{1}{K} \sum_{k=1}^{K} \| \mathbf{e}_k - \hat{\mathbf{e}}_s \|^2 \Big). \label{eq:6}
\end{equation}
The resulting codebook $\hat{\mathbf{e}}_s$ initializes subsequent fine-tuning.\vspace{-0pt}

\subsection{Learning Objective}\label{learningobj}
\noindent \textbf{Training Strategy.}~In addition to $\mathcal{L}_{\mathtt{VQ}}$, we integrate VQGAN, an enhanced version of the original VQ-VAE, to incorporate adversarial loss and perceptual loss into the proposed approach. The perceptual loss measures feature-space differences between original and reconstructed CSI, while the keypoint loss directly supervises pose estimation \cite{26}. Specifically, we replace the CSI reconstruction term in Eq. (\ref{eq:3}) with perceptual loss \cite{34} and introduce a patch-based discriminator $D$, which uses adversarial training to differentiate between real and reconstructed CSI data, leading to
\begin{equation}
\mathcal{L}_{\mathtt{GAN}}(\{E, G, \hat{e}_s\}, D) = \big[\log D(\textbf{X}) + \log(1 - D(\hat{\textbf{X}}))\big]. \label{eq:7}
\end{equation}
We supervise HPE by minimizing the difference between predicted joint coordinates $\hat{\mathbf{Y}}=E_s(\hat{\mathbf{Z}}_q)$ and ground-truth coordinates $\mathbf{Y}$ corresponding to the input CSI:
\begin{equation}
\mathcal{L}_{\mathtt{keypoint}}=\|E_{s}(\hat{\textbf{Z}}_q)-\textbf{\ensuremath{\textbf{Y}}}\| ^2_2=\|\hat{\textbf{Y}}-\textbf{Y}\|^2_2. \label{eq:8}
\end{equation}
The keypoint loss links representation learning to HPE by penalizing errors in the predicted joint coordinates. Together with the VQ and adversarial terms, it trains the latent representation to preserve CSI structure and pose information.

\vspace{0pt}

\noindent \textbf{Algorithm summary.}~The objective is to find the optimal compression model $\mathcal{O}^* = \{E^*, G^*, E_{s}^*, \hat{\textbf{e}}_s^*\}$:
\begin{align}
\mathcal{O}^* = & \, \arg \min_{E, G, \hat{\textbf{e}}_s} \max_{D} \mathbb{E}_{x \sim p(x)} \Big[  \mathcal{L}_\mathtt{VQ}(E, G, \hat{\textbf{e}}_s) \nonumber\\
& + \lambda \mathcal{L}_\mathtt{GAN}(\{E, G, \hat{\textbf{e}}_s\}, D) + \mathcal{L}_{\mathtt{keypoint}}(E_{s}) \Big] 
\label{eq:loss}
\end{align}
where $\lambda$ is calculated as
\begin{equation}
\lambda = \frac{\nabla_{G_{L}} [\mathcal{L}_\mathtt{rec}]}{\nabla_{G_{L}} [\mathcal{L}_\mathtt{GAN}] + \delta}\label{eq:10}
\end{equation}
with $\mathcal{L}_\mathtt{rec}$ and $\nabla_{G_{L}}[\cdot]$ being the perceptual loss \cite{58} and the gradient with respect to the decoder's last layer, respectively. $\delta = 10^{-6}$ is used to maintain numerical stability. Specifically, we illustrate the $\mathtt{TinySense}$ training and deployment in Algorithm \ref{alg1}. Using existing CSI data, we first train the $\mathtt{TinySense}$ model offline, then deploy the feature extractor $E$ at the local device, while the cloud server hosts the decoder $G$ and Estimator $E_s$. The trained CSI codebook is stored on local devices and the server for compression and reconstruction.\vspace{0pt}

\begin{algorithm}[t]
  \caption{Training and Deployment of $\mathtt{TinySense}$}
\label{alg1}
\textbf{Input:} CSI samples $\mathbf{X}$ and pose coordinates $\mathbf{Y}$, model weights \(\theta_E\), \(\theta_G\), \(\theta_{E_{s}}\), CSI codebook \(\textbf{e}_k\), and number of iterations \(N_{\text{iter}}\).

\textbf{Initialize:} Encoder \(E\), decoder \(G\), and estimator \(E_s\) with parameters \(\theta_E\), \(\theta_G\), and \(\theta_{E_{s}}\), respectively.

\textbf{While} \(i \leq N_{\text{iter}}\) \textbf{do}
\begin{enumerate}
    \item Input the labelled data \((\textbf{X}, \textbf{Y})\);
    \item Extract features \(\textbf{Z}\) from \(\textbf{X}\) using the encoder \(E(\textbf{X})\). Map \(\textbf{Z}\) to  VQ-indices map \(\textbf{I}\) using codebook \(\textbf{e}_k\), and then retrieve quantized features \(\textbf{Z}_q\);
    \item Compress \(\textbf{I}\) into a binary bitstream and transmit to the cloud;
    \item Decompress the bitstream on the cloud side to retrieve  the VQ-indices map \(\hat{\textbf{I}}\) by  Binary Decoder;
    \item Restore the quantized feature \(\textbf{Z}_q\) from \(\hat{\textbf{I}}\);
    \item Reconstruct CSI data \(\hat{\textbf{X}}\) using decoder \(G\) and predict results with  estimator \(E_s\);
    \item Update encoder \(\theta_E\) and new  codebook \(\textbf{e}_k \mathrel{:=} \textbf{e}_s\) by solving  (\ref{eq:3})  with K-means clustering;
    \item Update \(\theta_{E_{s}}\) and \(\theta_G\) by minimizing (\ref{eq:7}) and (\ref{eq:8}).
\end{enumerate}
$\triangleright$ \textbf{Deploy} $E$ at edge side, $G$ and $E_s$ at cloud server;

$\triangleright$ \textbf{Store} the codebook $\textbf{e}_s$ on both sides.
\end{algorithm} 

\subsection{Lost VQ-indices Prediction}\label{sec:LostVQ}
The recovery module predicts unavailable VQ indices before CSI reconstruction and HPE. It operates on an index map with known missing positions. The evaluation in Section~\ref{sec:experimental-result} tests synthetic index erasure.\vspace{2pt}

\begin{figure}[ht]
  \centering
  \includegraphics[width=0.8\linewidth]{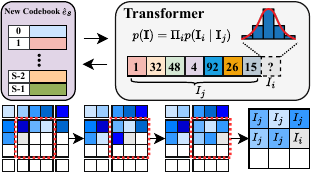}
  \caption{Recovery Transformer for completing missing VQ indices.} 
  
  \label{fig:figure3}
\end{figure}

\noindent \textbf{VQ-indices Transformer.}~As discussed in Sec. \ref{sec:VQ-compression}, the input CSI data $\textbf{X}$ can be represented by a map of the VQ indices $\textbf{I}$, derived from its encodings. Specifically, the quantized representation of the CSI data $\textbf{X}$ is defined as $\textbf{Z}_q = \textbf{q}(E(\textbf{X})) \in \mathbb{R}^{\frac{F}{M} \times \frac{T}{M}  \times D}$, which corresponds to a sequence of indices $\textbf{I\ensuremath{\in(1,..., S)}}^{\frac{F}{M} \times \frac{T}{M} }$ from the codebook. Each index is obtained by substituting the code with its corresponding index in the codebook $\textbf{e}_s$ as
\begin{equation}
    \textbf{I}_{ij} = k,\,\,  \text{such that}\,\,  (\textbf{Z}_q)_{ij} = \textbf{Z}_k. \label{eq:11}
\end{equation}
By mapping indices back to their respective entries in the codebook, the quantized features $\hat{\textbf{Z}}_q=\hat{\textbf{Z}}_{e_{ij}}$ can be easily reconstructed, and the CSI data is decoded as $\hat{\textbf{X}} = G(\hat{\textbf{Z}}_q)$.

{The base $\mathtt{TinySense}$ design uses a masked, bidirectional recovery Transformer, denoted by $T_{\mathrm{rec}}$. During training, a random subset $\mathcal{M}$ of VQ indices is replaced by a special \texttt{[mask]} token. The model predicts those indices from the visible context $\mathbf{I}_{\bar{\mathcal{M}}}$ on both sides of each masked position by minimizing}
\begin{equation}
{
\mathcal{L}_{\mathtt{rec}}
=
\mathbb{E}_{\mathbf{I},\mathcal{M}}
    \Big[-\sum_{i\in\mathcal{M}}
\log p_{T_{\mathrm{rec}}}
\bigl(\mathbf{I}_i\mid\mathbf{I}_{\bar{\mathcal{M}}}\bigr)
\Big]}
\label{eq:12}
\end{equation}
where $\mathcal{L}_{\mathtt{rec}}$ denotes the index-recovery loss, distinct from the perceptual loss with the same notation in Eq.~\eqref{eq:10}. The mask ratio $\alpha\in[0,1]$ determines the fraction of unavailable indices, so $|\mathcal{M}|=\alpha N$. At inference, $T_{\mathrm{rec}}$ predicts each unavailable position before CSI reconstruction. Because it can use observations on both sides of a position, $T_{\mathrm{rec}}$ is different from the strictly causal entropy Transformer $q_{\psi}$ introduced for $\mathtt{TinySense+}$ in Sec.~\ref{sec:entropy_model}.\vspace{3pt}

\noindent \textbf{Sliding window.} To bound the input length of $T_{\mathrm{rec}}$, we center a $P\times P$ window on each unavailable index $\mathbf{I}_i$, as illustrated in Fig.~\ref{fig:figure3}. The model may use any visible indices in that window, including context before and after position $i$. This bidirectional context is appropriate for arbitrary erasure recovery but cannot be used to generate the causal probabilities required by arithmetic decoding. The completed index map $\mathbf{I}_{\mathrm{predict}}$ is mapped through the shared codebook and decoder $G$ to reconstruct $\hat{\mathbf{X}}$.\vspace{-0pt}

\begin{figure}[tb]
  \centering
  \includegraphics[width=0.9\linewidth]{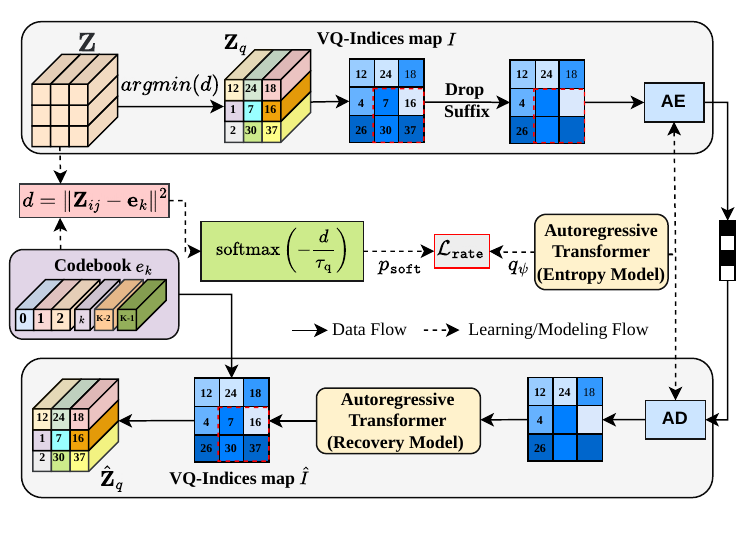}
\caption{{The $\mathtt{TinySense+}$ pipeline: causal entropy modeling for arithmetic encoding (AE) and decoding (AD), followed by prefix-conditioned suffix completion using the same model.}}
\vspace{-0.5cm}
\label{fig:tinysense_plus_pipeline}
\end{figure}

\section{TinySense+: Entropy-Aware Dynamic Compression}
\label{sec:TinySense+}

{The base $\mathtt{TinySense}$ framework reduces CSI traffic by transmitting VQ indices, but its fixed-length representation leaves statistical redundancy in the bitstream. For a codebook of size $K$, every index requires $\lceil\log_2 K\rceil$ bits, irrespective of its frequency or context. It therefore cannot exploit a non-uniform index distribution, and changing the K-means-reduced codebook size provides only discrete operating rates.}

{To address these limitations, we introduce $\mathtt{TinySense+}$ (Fig.~\ref{fig:tinysense_plus_pipeline}), an entropy-aware framework with three coupled components. First, a strictly causal Transformer $q_{\psi}$ estimates each index distribution from previously decoded indices, allowing an arithmetic coder to represent predictable sequences with fewer bits. Second, a differentiable rate--distortion term exposes the expected code length to the encoder during training. Third, PCLC reuses $q_{\psi}$ to generate an intentionally omitted suffix from a transmitted prefix. The prefix ratio then controls the communication--accuracy trade-off at test time without retraining or codebook resizing. The causal $q_{\psi}$ is separate from the masked, bidirectional $T_{\mathrm{rec}}$ used by the base $\mathtt{TinySense}$ design.}

\subsection{Entropy Model}\label{sec:entropy_model}
The core component of $\mathtt{TinySense+}$ is a causal autoregressive entropy model that estimates the probability distribution of the VQ-index sequence produced by the encoder. After the CSI input $\mathbf{X}$ is encoded and quantized, the resulting VQ-index map is flattened into a sequence as
  \begin{equation}
      \mathbf{I} = (i_1, i_2, \ldots, i_N), \quad \text{with}\,\, i_t \in \{1,2,\ldots,K\}
  \end{equation}
  {where $N$ is the number of latent tokens and $K$ is the active codebook size. In this section, $K$ also covers a reduced codebook, whose size is denoted by $S$ in Section~\ref{sec:VQ-compression}.}

  To exploit this statistical structure, $\mathtt{TinySense+}$ learns an entropy model $q_{\psi}$ parameterized by an autoregressive Transformer. For each position $t$, the model predicts a categorical distribution over the codebook conditioned on the available context $c_t$ as
  \begin{equation}
      q_{\psi}(i_t = k \mid c_t), \quad k \in \{1,2,\ldots,K\}.
  \end{equation}
  Since the entropy model is autoregressive, the context is causal, $c_t = (i_1, i_2, \ldots, i_{t-1}) = i_{<t}$, i.e., only the
previously coded indices. This constraint is necessary because arithmetic decoding must reproduce exactly the same probability
distribution used during encoding, and at decoding time only the already-decoded indices are available.

  The entropy model is trained to maximize the likelihood of the observed VQ-index sequence, or equivalently to minimize the
cross-entropy loss
  \begin{equation}
      \mathcal{L}_{\mathtt{ent}}
      =
      \mathbb{E}_{\mathbf{I}}
      \Big[
      - \sum_{t=1}^{N} \log q_{\psi}(i_t \mid c_t)
      \Big].
      \label{eq:entropy_model_loss}
  \end{equation}
  The ideal code length of index $i_t$ is approximately $-\log_2 q_{\psi}(i_t \mid c_t)$ bits, so accurate conditional probabilities improve coding efficiency.

The causal constraint separates $q_{\psi}$ from $T_{\mathrm{rec}}$ in Sec.~\ref{sec:LostVQ}. The base model recovers arbitrary masked locations using visible context on both sides, while arithmetic decoding, by contrast, can condition only on indices already decoded. Hence $q_{\psi}$ uses a causal mask and can also generate the PCLC suffix sequentially from the received prefix.

\subsection{Differentiable Rate-Distortion Objective}\label{sec:rd_objective}

 Entropy coding benefits from predictable VQ-index sequences. However, the base $\mathtt{TinySense}$ model is trained mainly to preserve CSI reconstruction quality and
HPE accuracy. It does not explicitly encourage the encoder to produce low-entropy or easily compressible index sequences. To address
this issue, $\mathtt{TinySense+}$ introduces a differentiable rate-distortion objective that directly regularizes the encoder toward
entropy-aware representations.

  The main challenge is that the standard VQ operation relies on a hard nearest-neighbor assignment
  \begin{equation}
      i_t = \arg\min_{k \in \{1,\ldots,K\}} \| \mathbf{z}_t - \mathbf{e}_k \|_2^2
  \end{equation}
  where $\mathbf{z}_t$ is the encoder output at latent position $t$, and $\mathbf{e}_k$ is the $k$-th codeword in the codebook. This
hard assignment is non-differentiable, so a rate loss computed on the discrete index $i_t$ cannot directly update the encoder.

To make the rate term differentiable, we replace the hard assignment with a distance-aware soft assignment during training. Specifically, for each latent vector $\mathbf{z}_t$, we compute a probability distribution over the codebook as
\begin{equation}
p_{\theta}(k\mid\mathbf{z}_t)=
\frac{\exp\!\left(-\|\mathbf{z}_t-\mathbf{e}_k\|_2^2/
\tau_{\mathrm{q}}\right)}
{\sum_{m=1}^{K}\exp\!\left(
-\|\mathbf{z}_t-\mathbf{e}_m\|_2^2/\tau_{\mathrm{q}}\right)}
\end{equation}
where $\tau_{\mathrm{q}}>0$ is the soft-assignment temperature. A smaller $\tau_{\mathrm{q}}$ concentrates probability on the nearest codewords, whereas a larger value produces a more diffuse distribution.

Using this soft assignment, we define the expected code length of the latent sequence under the entropy model
  \begin{equation}
      \mathcal{L}_{\mathtt{rate}}
      =
      \mathbb{E}
      \Big[
      \sum_{t=1}^{N}
      \sum_{k=1}^{K}
      p_{\theta}(k \mid \mathbf{z}_t)
      \left(
      - \log_2 q_{\psi}(k \mid c_t)
      \right)
      \Big].
      \label{eq:rate_loss}
  \end{equation}
  This loss encourages the encoder to produce indices that are predictable under the entropy model and therefore require fewer bits.

  {The encoder, codebook, decoder, and HPE estimator retain the original $\mathtt{TinySense}$ objective and add the differentiable rate term:}
  \begin{equation}
      {\mathcal{L}_{\mathtt{TinySense+}}
      =
      \mathcal{L}_{\mathtt{TinySense}}
      +
      \gamma \mathcal{L}_{\mathtt{rate}}}.
      \label{eq:tinysense_plus_loss}
  \end{equation}

The rate term encourages predictable index sequences, while the keypoint term penalizes pose errors. Their combination balances the expected coding cost with the pose information retained in the learned representation. The entropy Transformer is updated with $\mathcal{L}_{\mathtt{ent}}$ in Eq.~\eqref{eq:entropy_model_loss}. During an encoder update, $q_{\psi}$ is held fixed, so $\mathcal{L}_{\mathtt{rate}}$ reshapes the encoder-induced distribution instead of being reduced only by changing the entropy model. The two updates alternate. The coefficient $\gamma$ controls the measured trade-off: $\gamma=0$ removes the rate penalty while retaining entropy-coded transmission, whereas a larger $\gamma$ favors more predictable indices and may trade reconstruction or HPE accuracy for a shorter coded stream.

\subsection{AC Index Transmission}
\label{sec:ac_transmission}

  After training, $\mathtt{TinySense+}$ uses the hard VQ assignment for inference, producing a discrete index sequence
  \begin{equation}
      \mathbf{I} = (i_1, i_2, \ldots, i_N).
  \end{equation}
  Unlike the base $\mathtt{TinySense}$ framework, which represents each index using a fixed number of bits, $\mathtt{TinySense+}$
transmits this sequence using arithmetic coding guided by the autoregressive entropy model. For each index $i_t$, the entropy model estimates the conditional probability distribution $q_{\psi}(i_t \mid c_t)$ from the valid
coding context $c_t$. The arithmetic encoder then uses this probability distribution to encode $i_t$ into a compact variable-length
bitstream. The expected bit cost of the transmitted sequence can be approximated as
  \begin{equation}
      B_{\mathtt{AC}}
      \approx
      \sum_{t=1}^{N}
      -\log_2 q_{\psi}(i_t \mid c_t).
      \label{eq:ac_bits}
  \end{equation}
  In contrast, fixed-length coding requires $B_{\mathtt{fixed}}      =
      N \lceil \log_2 K \rceil$
  bits for a sequence of length $N$ and codebook size $K$. Therefore, the gain of $\mathtt{TinySense+}$ comes from exploiting the
non-uniform and context-dependent structure of the learned VQ-index sequence. At the receiver side, the cloud server runs the same autoregressive entropy model to reproduce the probability distribution
used by the encoder. Arithmetic decoding then reconstructs the exact VQ-index sequence $\mathbf{I}$ from the received bitstream. The
recovered indices are mapped back to the corresponding codebook vectors, forming the quantized latent representation
$\hat{\mathbf{Z}}_q$, which is then processed by the decoder and HPE estimator as in the base $\mathtt{TinySense}$ framework.

Arithmetic coding is lossless with respect to the VQ-index sequence. Consequently, when the full sequence is transmitted, any reconstruction change relative to the base system comes from the learned VQ representation and rate--distortion training, not from the arithmetic coder.

Eq.~\eqref{eq:ac_bits} approximates code length from model probabilities, while finite-precision coding and stream termination can affect the realized length. The compression ratio is computed from the actual arithmetic-coded bitstream length $\eta_{\mathtt{TinySense+}} = B_{\mathtt{raw}}/B_{\mathtt{AC}}$, where $B_{\mathtt{raw}}$ denotes the bit cost of transmitting the original CSI data. Since $B_{\mathtt{AC}}$ depends on the learned
probability distribution rather than only the codebook size, $\mathtt{TinySense+}$ provides more fine-grained compression behavior
than fixed-length VQ coding.

 \subsection{Test-Time Rate Adjustment via PCLC}
  \label{sec:prefix_transmission}

Beyond entropy-coded transmission, $\mathtt{TinySense+}$ also introduces PCLC, enabling test-time rate adjustment through prefix transmission. This mechanism is designed for highly dynamic network conditions, where the available bandwidth may suddenly decrease and the edge device
cannot afford to transmit the complete VQ-index sequence. Given the VQ-index sequence $\mathbf{I} = (i_1, i_2, \ldots, i_N)$, the edge device transmits only the first $M$ indices, $\mathbf{I}_{\mathtt{pre}} = (i_1, i_2, \ldots, i_M),
      \quad M = \lfloor \rho N \rfloor$,
where $M$ denotes the prefix length in this subsection, and $\rho \in (0,1]$ is the prefix ratio determined by the available bandwidth. A smaller $\rho$ reduces the transmitted payload,
while a larger $\rho$ preserves more original latent information. The missing suffix is defined as $\mathbf{I}_{\mathtt{suf}} = (i_{M+1}, i_{M+2}, \ldots, i_N)$. After receiving $\mathbf{I}_{\mathtt{pre}}$, the server uses the same autoregressive entropy model $q_{\psi}$ to predict the omitted suffix conditioned on
the transmitted prefix:
  \begin{equation}
      \hat{\mathbf{I}}_{\mathtt{suf}}
      =
      \arg\max_{\mathbf{I}_{\mathtt{suf}}}
      \prod_{t=M+1}^{N} q_{\psi}(i_t \mid i_{<t}).
      \label{eq:prefix_recovery}
  \end{equation}
  In practice, the suffix is decoded greedily, one index at a time, conditioned on the prefix and previously generated indices. Greedy decoding approximates the sequence-level objective in Eq.~\eqref{eq:prefix_recovery}. We note that it does not guarantee the global optimum. The complete index sequence is then reconstructed as $ \hat{\mathbf{I}}     =
      (\mathbf{I}_{\mathtt{pre}}, \hat{\mathbf{I}}_{\mathtt{suf}}). $
  The completed sequence $\hat{\mathbf{I}}$, whose predicted suffix may differ from the original indices, is mapped back to the codebook vectors and passed to the decoder and HPE estimator. With prefix transmission, the effective compression rate becomes
  \begin{equation}
      \eta_{\mathtt{pre}}
      =
      \frac{B_{\mathtt{raw}}}{B_{\mathtt{pre}}}
  \end{equation}
  where $B_{\mathtt{pre}}$ is the bit cost of transmitting the prefix sequence. If arithmetic coding is applied to the prefix, then
  \begin{equation}
      B_{\mathtt{pre}}
      \approx
      \sum_{t=1}^{M}
      -\log_2 q_{\psi}(i_t \mid c_t).
  \end{equation}
  Thus, $\mathtt{TinySense+}$ can adapt to bandwidth constraints in two ways: by reducing the entropy-coded bit cost of each
transmitted index and by reducing the number of transmitted indices through prefix transmission.

  We recall that reducing $\rho$ increases compression but requires a larger predicted suffix, potentially reducing CSI reconstruction and HPE accuracy. This trade-off is adjusted at test time without retraining the encoder or resizing the codebook.

\section{Experimental Evaluation} \label{sec:experimental-result}
\begin{table*}[t]
    \centering
    \small
      \caption{Compression Performance Comparison across Different Schemes on both MM-Fi and WiPose Datasets (\textit{``N/A" indicates methods not applicable to the task})}%
    \begin{tabular}{c|c|cccc|cccc}
\hline
\multirow{2}{*}{\textbf{Method}} & \multirow{2}{*}{$\boldsymbol{\eta}$} & \multicolumn{4}{c|}{\textbf{MM-Fi}} & \multicolumn{4}{c}{\textbf{WiPose}} \\
\cline{3-10}
 & & \textbf{NMSE (dB)} $\downarrow$ & $\textbf{PCK}_{30} \uparrow$ & $\textbf{PCK}_{20} \uparrow$ & \textbf{MPJPE} $\downarrow$ & \textbf{NMSE (dB)} $\downarrow$ & $\textbf{PCK}_{10} \uparrow$ & $\textbf{PCK}_{5} \uparrow$ & \textbf{MPJPE} $\downarrow$ \\
\hline
LASSO & 4 & -8.42 & N/A & N/A & N/A & -15.64 & N/A & N/A & N/A \\
DeepCMC & 4 & -9.58 & N/A & N/A & N/A & -16.86 & N/A & N/A & N/A \\
GLC & 4 & -13.18 & N/A & N/A & N/A & -20.75 & N/A & N/A & N/A \\
\hline
\multirow{4}{*}{CsiNet} 
 & 4 & -4.75 & \multirow{4}{*}{N/A} & \multirow{4}{*}{N/A} & \multirow{4}{*}{N/A} & -8.16 & \multirow{4}{*}{N/A} & \multirow{4}{*}{N/A} & \multirow{4}{*}{N/A} \\
 & 16 & -4.25 &  &  &  & -7.93 &  &  &  \\
 & 64 & -4.05 &  &  &  & -7.41 &  &  &  \\
 & 128 & -3.72 &  &  &  & -6.98 &  &  &  \\
\hline
\multirow{4}{*}{\ph{RSCNet}}
 & 4  & -13.02 & 55.14 & 42.62 & 170.15 & -14.16 & 42.48 & 28.14 & 57.89 \\
 & 16 & -12.53 & 52.26 & 40.15 & 177.26 & -14.06 & 37.12 & 23.69 & 64.13 \\
 & 32 & -12.36 & 49.24 & 36.72 & 186.18 & -13.48 & 32.54 & 18.14 & 75.16 \\
 & 64 & -11.17 & 46.14 & 31.98 & 189.74 & -13.35 & 23.17 & 12.68 & 88.17 \\
\hline
\multirow{5}{*}{EfficientFi}
 & 4 & -14.94 & 62.15 & 48.01  & 166.62 & -18.17 & 62.14 & 42.16 & 34.62 \\
 & 16 & -13.84 & 57.68 & 46.15 & 172.14 & -17.58 & 59.73 & 38.96 & 40.18 \\
 & 86 & -13.26 & 52.47 & 40.68 & 178.18 & -14.24 & 54.12 & 33.14 & 48.87 \\
 & 570 & -11.45 & 46.24 & 36.16 & 185.28 & -11.36 & 46.95 & 26.89 & 62.14 \\
 & 1710 & -9.28 & 42.28 & 25.62 & 207.78 & -7.07 & 34.18 & 15.39 & 86.16 \\
\hline
\multirow{6}{*}{\textbf{TinySense}}
 & 4 & -15.14 & 66.16 & 51.93 & 156.72 & -23.16 & 79.45 & 67.62 & 16.72 \\
 & 16 & -14.82 & 64.24 & 48.59 & 163.21 & -22.56 & 76.19 & 64.18 & 19.16 \\
 & 86 & -14.42 & 63.15 & 46.51 & 166.87 & -22.12 & 70.82 & 59.89 & 25.47 \\
 & 215 & -13.74 & 60.89 & 41.21 & 170.98 & -21.48 & 63.16 & 54.46 & 30.96 \\
 & 570 & -13.17 & 56.04 & 37.75 & 177.26 & -21.01 & 54.96 & 45.39 & 38.42 \\
 & 1710 & -10.72 & 50.85 & 32.93 & 189.67 & -18.16 & 42.19 & 36.72 & 50.58 \\
\hline
\multirow{6}{*}{\textbf{TinySense+}}
 & 20   & -14.47 & 66.36 & 49.81 & 156.15 & -21.22 & 77.53 & 66.05 & 17.85 \\
 & 38   & -14.01 & 64.50 & 48.34 & 160.32 & -20.49 & 75.40 & 63.50 & 21.10 \\
 & 300  & -11.66 & 61.34 & 44.08 & 170.26 & -18.82 & 68.20 & 57.15 & 27.80 \\
 & 650  & -10.63 & 61.12 & 43.96 & 171.18 & -17.12 & 59.79 & 52.79 & 35.67 \\
 & 1630 & -10.25 & 57.89 & 40.77 & 178.21 & -16.68 & 52.17 & 43.12 & 42.45 \\
 & 5840 & -9.01  & 51.80 & 37.92 & 191.84 & -15.85 & 40.36 & 32.15 & 59.20 \\
 \hline
\end{tabular}
          \label{tab:compressionTask} 
\end{table*}
\subsection{Experimental Setup}\label{sec:eval_setup}

\noindent\textbf{Datasets.} We use MM-Fi~\cite{yang2023mmfi} and WiPose~\cite{Toward2020}. MM-Fi includes CSI and pose annotations with $17$ joints from $40$ participants performing $27$ actions: $14$ daily activities and $13$ rehabilitation exercises. WiPose contains $18$-joint annotations and CSI from $12$ volunteers performing $12$ actions. The WiPose data are randomly divided into training and test sets.\vspace{3pt}

\noindent\textbf{Implementation.} Models are implemented in PyTorch and trained on an RTX 8000 for $50$ epochs with Adam, batch size $128$, and learning rate $0.001$. The embedding dimension is $D=256$. We evaluate active codebook sizes $K\in\{2,4,8,16,32,64,128,256,512,1024\}$, using K-means for reduced codebooks. The $\mathtt{TinySense+}$ settings are $\gamma=0.1$ and $\tau_{\mathrm{q}}=0.01$, except where explicitly varied. The reference training configuration reports $\lambda=0.5$ and an Adam momentum coefficient of $0.9$.\vspace{3pt}

\noindent\textbf{Baselines.} Compression comparisons include the least absolute shrinkage and selection operator (LASSO)~\cite{29}, DeepCMC~\cite{31}, CsiNet~\cite{30}, GLC~\cite{GLC}, RSCNet~\cite{20}, and EfficientFi~\cite{19}. HPE comparisons include MetaFi++~\cite{MetaFi++}, PerUnet~\cite{Wipose}, WiSPPN~\cite{Wang2019CanWE}, PiW-3D~\cite{PiW3D}, and AdaPose~\cite{AdaPose}. The code repository is given in Abstract.\vspace{3pt}

\noindent\textbf{Metrics and coding cost.} We report normalized mean square error (NMSE, dB) for CSI reconstruction, percentage of correct keypoints (PCK, \%), and mean per-joint position error (MPJPE) for HPE~\cite{PCK}. Lower NMSE and MPJPE and higher PCK indicate better performance. Compression ratio is $\eta=B_{\mathtt{raw}}/B_{\mathtt{coded}}$, with both quantities expressed in bits. The raw reference uses $32$ bits per amplitude sample; it does not include a second phase component. The fixed-length payload is $N\lceil\log_2 K\rceil$ bits. For $\mathtt{TinySense+}$, the denominator is the actual arithmetic-coded byte length multiplied by eight.\vspace{3pt}

\begin{figure}[t]
\centering
\includegraphics[width=\linewidth]{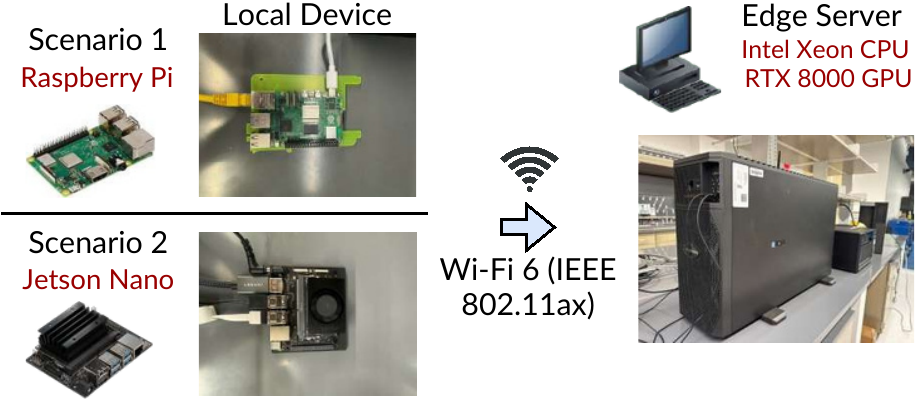}
\caption{Testbed for measuring edge--server latency, local energy, and transmitted payload.}
\label{fig:setup}
\end{figure}

\noindent\textbf{Testbed.} Scenario~1 uses a Raspberry Pi~5 and a Linux server with an NVIDIA Quadro RTX~8000; Scenario~2 replaces the edge device with a Jetson Nano (Fig.~\ref{fig:setup}). Double buffering overlaps processing of one CSI window with acquisition of the next, using a one-second acquisition window. Compressed data are sent through the Transmission Control Protocol/Internet Protocol (TCP/IP). The pipeline latency covers encoding, transmission, and decoding after a window becomes available. This processing latency is distinct from the acquisition duration and the age of the samples used for a pose estimate. The index-erasure experiments below are separate from TCP transport behavior.

\subsection{Compression and Pose Estimation Performance}\label{sec:eval_performance}

\noindent\textbf{Compression task.} Table~\ref{tab:compressionTask} reports NMSE, PCK, and MPJPE at matched or nearby compression ratios. At $\eta=4$, $\mathtt{TinySense}$ provides lower reconstruction error than the listed reconstruction-oriented baselines on both datasets. The pose comparisons apply only to methods with the HPE results. At the most compressed operating point shared with EfficientFi, $\eta=1710$, it obtains a PCK$_{20}$ of $32.93\%$ on MM-Fi, compared with $25.62\%$ for EfficientFi. These comparisons show that the VQ representation preserves information useful for HPE even when the reconstructed CSI is strongly compressed.
$\mathtt{TinySense+}$ extends the evaluated operating range through rate-aware training and arithmetic coding. The lower block of Table~\ref{tab:compressionTask} spans $\eta=20$ to $\eta=5840$. At $\eta=5840$, it reports a PCK$_{20}$ of $37.92\%$ and an MPJPE of $191.84$ on MM-Fi. This operating point characterizes the compression--accuracy trade-off on the evaluated dataset.\vspace{3pt}
\begin{table}[ht!]
    \centering
     \caption{HPE Performance on MM-Fi and WiPose (Parameter counts are in millions (M); N/A indicates an unreported compression ratio)} 
    \small
\begin{tabular}{ccccc}
\hline 
\multicolumn{5}{c}{$\textbf{MM-Fi}$}\tabularnewline
\hline 
$\textbf{Method}$ & $\eta$ &  $\textbf{PCK}_{20} \uparrow$ & $\textbf{MPJPE\ensuremath{}} \downarrow$ & $\textbf{Params} \downarrow$\tabularnewline
\hline 
WiSPPN  & N/A  & 45.41 & 166.59 & 26.78M\tabularnewline
PerUnet & N/A &  52.12 & 154.66 & 34.51M\tabularnewline
MetaFi++  & N/A  & 48.64 & 162.45 & 26.42M\tabularnewline
PiW-3D  & N/A    & 52.18  & 155.69 & 16.47M \tabularnewline
AdaPose & N/A   & 50.58  & 161.75 & 14.62M\tabularnewline
\hline 
\multirow{3}{*}{\textbf{TinySense}} & 2  &  54.12 & 152.16  &\multirow{3}{*}{1.66M} \tabularnewline
 & 4  & 51.93 & 156.72 & \tabularnewline
 & 16  & 48.59 & 163.21 & \tabularnewline
\hline
\multirow{3}{*}{\textbf{TinySense+}} & 20  &  49.81 & 156.15  &\multirow{3}{*}{1.66M} \tabularnewline
 & 38  & 48.34 & 160.32 & \tabularnewline
 & 300  & 44.08 & 170.26 & \tabularnewline

\hline
\multicolumn{5}{c}{\textbf{WiPose}}\tabularnewline
\hline 
$\textbf{Method}$ & $\eta$ &  $\textbf{PCK}_{5} \uparrow$ & $\textbf{MPJPE\ensuremath{}} \downarrow$ & $\textbf{Params} \downarrow$\tabularnewline
\hline 
WiSPPN   & N/A &  52.95 & 32.37 & 26.33M\tabularnewline
PerUnet   & N/A & 59.94 & 25.12 & 33.85M\tabularnewline
MetaFi++ & N/A &  53.64 & 30.62 & 25.58M\tabularnewline
PiW-3D  & N/A   & 58.45  & 27.02  & 18.63M \tabularnewline
AdaPose  & N/A  & 55.12  & 28.84 & 16.25M \tabularnewline
\hline 
\multirow{3}{*}{\textbf{TinySense}} & 8 &  65.36 & 17.75 & \multirow{3}{*}{7.36M} \tabularnewline
 & 32  & 63.89 & 21.47 & \tabularnewline
& 64  & 60.12 & 24.16 &   \tabularnewline
\hline 
\multirow{3}{*}{\textbf{TinySense+}} & 20 & 66.05 & 17.85 & \multirow{3}{*}{14.82M} \tabularnewline
 & 38  & 63.50 & 21.10  & \tabularnewline
 & 300 & 57.15 & 27.80 & \tabularnewline
\hline 
\end{tabular}$ $
 \label{tab:HPEtasks}
\end{table}

\noindent \textbf{HPE tasks.} Table~\ref{tab:HPEtasks} compares compressed sensing pipelines with HPE methods that do not report a compression rate. On MM-Fi, $\mathtt{TinySense}$ reaches a PCK$_{20}$ of $54.12\%$ at $\eta=2$ and $48.59\%$ at $\eta=16$, showing the accuracy cost of a smaller payload. $\mathtt{TinySense+}$ shifts this trade-off toward larger compression ratios: at $\eta=300$, it attains a PCK$_{20}$ of $44.08\%$ and an MPJPE of $170.26$. On WiPose, it reaches a PCK$_5$ of $66.05\%$ at $\eta=20$. Parameter counts are reported by dataset and configuration. \vspace{3pt}

\begin{table}[ht!]
    \centering
    \small
    \caption{Effect of the recovery Transformer under synthetic index erasure on MM-Fi}%
    \begin{tabular}{@{\hskip 3pt}c@{\hskip 6.5pt}|@{\hskip 6.5pt}c@{\hskip 6.5pt}c@{\hskip 6.5pt}c@{\hskip 6.5pt}|c@{\hskip 6.5pt}c@{\hskip 6.5pt}c@{\hskip 3pt}}\hline
 & \multicolumn{3}{c|}{\textbf{$\mathtt{TinySense}$}} & \multicolumn{3}{c}{$\begin{array}{c}
\textbf{$\mathtt{TinySense}$}\\
\textbf{Without Transformer}
\end{array}$}\tabularnewline
\hline 
$\epsilon$ & $\textbf{NMSE\ensuremath{\downarrow}}$ & $\textbf{PCK}_{20}$\ensuremath{\uparrow} & $\textbf{MPJPE\ensuremath{\downarrow}}$ & $\textbf{NMSE\ensuremath{}}$ & $\textbf{PCK}_{20}$ & $\textbf{MPJPE}$\tabularnewline
\hline 
0.1 & -12.02 & 39.16 & 171.45 & -10.14 & 31.64 & 190.26\tabularnewline
0.3 & -11.24 & 36.12 & 178.63 & -9.26 & 26.63 & 198.74\tabularnewline
0.5 & -10.88 & 32.69 & 186.54 & -8.56 & 21.16 & 218.65\tabularnewline
0.7 & -10.04 & 30.14 & 192.68 & -7.94 & 18.63 & 227.98\tabularnewline
0.9 & -9.65 & 26.89 & 198.86 & -7.01 & 14.23 & 242.12\tabularnewline
\hline 
\end{tabular}

\label{tab:LostIndex1}
\end{table}

\noindent \textbf{Synthetic VQ-index erasure.}~We isolate the recovery module by randomly removing a fraction $\epsilon$ of the VQ indices at a fixed compression rate of $215$ on MM-Fi. This is an index-erasure stress test rather than an end-to-end packet-loss experiment. Table~\ref{tab:LostIndex1} shows that performance decreases as more indices are removed, but $T_{\mathrm{rec}}$ consistently improves the reconstruction over the no-Transformer variant. At $\epsilon=0.9$, for example, it raises PCK$_{20}$ from $14.23\%$ to $26.89\%$ and reduces MPJPE from $242.12$ to $198.86$.\vspace{3pt}

\noindent\textbf{Operating points under index erasure.} Table~\ref{tab:LostIndex_Merged} reports configurations selected to keep PCK$_{20}>35\%$ as the index-erasure fraction $\epsilon$ increases. The base model's compression ratio decreases from $215$ to $16$. For $\mathtt{TinySense+}$, the ratio remains $1949$ for $\epsilon=0.1$--$0.5$, then decreases to $105$ and $20$ for $\epsilon=0.7$ and $0.9$.

At $\epsilon=0.9$, $\mathtt{TinySense+}$ gives higher PCK$_{20}$ ($36.08\%$ versus $35.01\%$) and lower NMSE ($-8.86$ versus $-5.29$~dB), but higher MPJPE ($204.23$ versus $157.03$). The comparison therefore shows a metric-dependent trade-off, not uniform superiority. These index-level results do not establish robustness of an arithmetic-coded stream to packet loss.

\begin{table}[ht!]
    \centering
    \caption{Adaptation Performance Comparison: Base $\mathtt{TinySense}$ vs. $\mathtt{TinySense+}$ under varying loss rates ($\epsilon$), aiming to maintain $\text{PCK}_{20} > 35\%$}
    \small
    \renewcommand{\arraystretch}{1.1}
\setlength{\tabcolsep}{5pt}
\begin{tabular}{c|c|c|ccc}
\hline 
\textbf{Model} & $\boldsymbol{\epsilon}$ & $\boldsymbol{\eta} \uparrow$ & $\textbf{NMSE} \downarrow$ & $\textbf{PCK}_{20}\uparrow$ & 
$\textbf{MPJPE} \downarrow$ \tabularnewline
\hline
\multirow{5}{*}{\begin{tabular}{@{}c@{}} \textbf{ $\mathtt{TinySense}$} \end{tabular}} 
& 0.1 & 215 & -6.29 & 36.51 & \textbf{148.54}  \tabularnewline
& 0.3 & 176 & -6.44 & 36.78 & \textbf{148.52} \tabularnewline
& 0.5 & 96  & -6.21 & 35.51 & \textbf{149.58} \tabularnewline
& 0.7 & 64  & -5.51 & 35.78 & \textbf{149.73} \tabularnewline
& 0.9 & 16  & -5.29 & 35.01 & \textbf{157.03} \tabularnewline
\hline
\multirow{5}{*}{\begin{tabular}{@{}c@{}} \textbf{$\mathtt{TinySense+}$}\end{tabular}} 
& 0.1 & \textbf{1949} & -8.05 & \textbf{46.09} & 165.53 \tabularnewline
& 0.3 & \textbf{1949} & -8.14 & \textbf{42.08} & 174.65 \tabularnewline
& 0.5 & \textbf{1949} & -7.95 & \textbf{37.86} & 185.32 \tabularnewline
& 0.7 & \textbf{105}  & \textbf{-8.83} & \textbf{37.08} & 187.53 \tabularnewline
& 0.9 & \textbf{20}   & \textbf{-8.86} & \textbf{36.08} & 204.23 \tabularnewline
\hline 
\end{tabular}
    \label{tab:LostIndex_Merged} 
\end{table}

\begin{figure*}[t]
    \centering
    \includegraphics[width=0.8\linewidth]
    {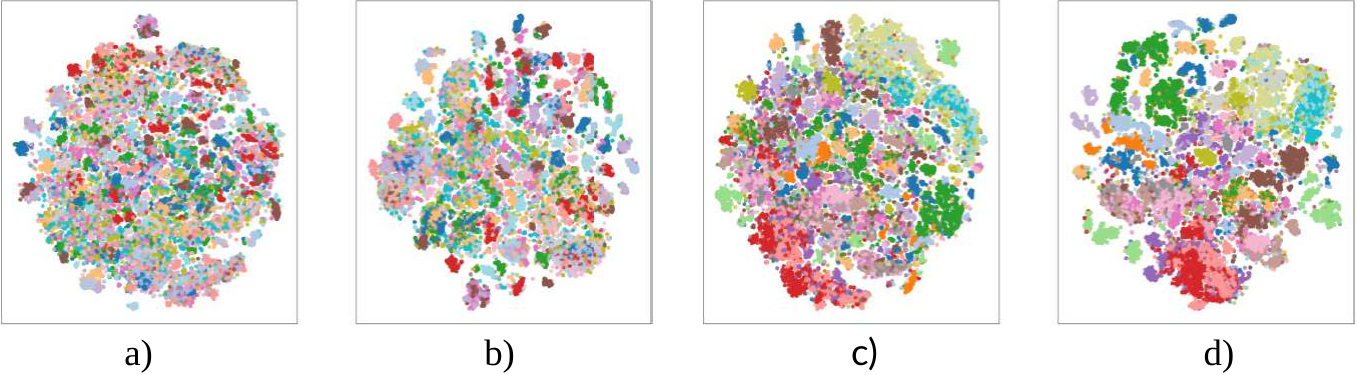}
    \caption{t-distributed stochastic neighbor embedding (t-SNE) visualizations of CSI representations: (a) raw CSI with action labels, (b) quantized features with action labels, (c) raw CSI with subject labels, and (d) quantized features with subject labels.}
    \label{fig:t-sne}
\end{figure*}
\begin{figure*}[t]
    \centering
    \includegraphics[width=0.95\linewidth]{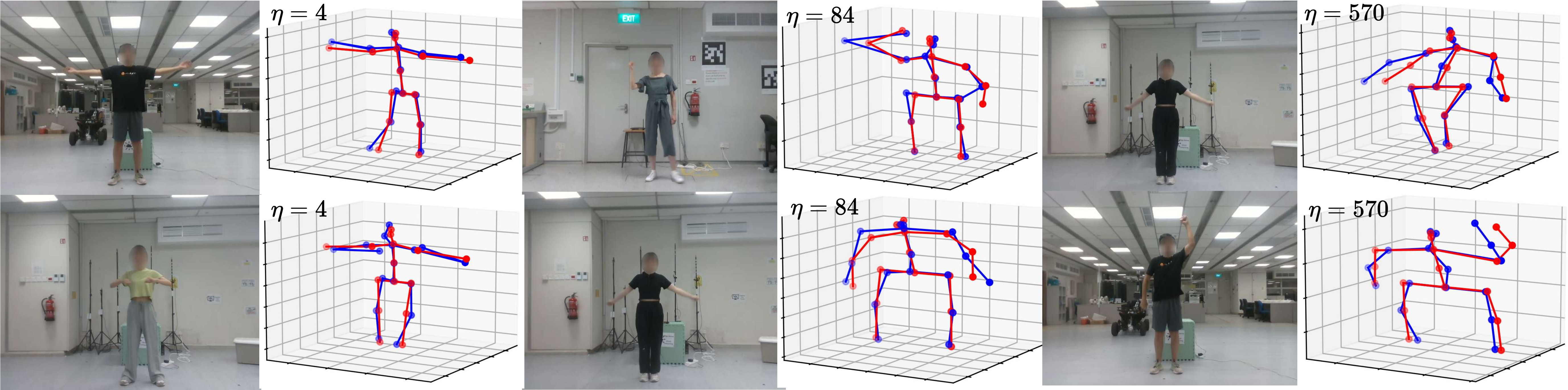}
    \caption{Visualization of the human pose landmarks generated by the vision model (red) and Wi-Fi model (blue) on MM-Fi at different compression ratios.} %
    \label{fig:keypoint_visualize}
\end{figure*}

 \noindent \textbf{Feature visualization.} We use t-distributed stochastic neighbor embedding (t-SNE) as a qualitative diagnostic of whether the compressed features retain action- and subject-related structure. In Fig.~\ref{fig:t-sne}(a), action clusters overlap, which is consistent with similar poses occurring in different actions. The quantized features in Fig.~\ref{fig:t-sne}(b) show greater separation, although the plot does not establish classification accuracy by itself. The views colored by subject identity in Fig.~\ref{fig:t-sne}(c)--(d) show a similar change after quantization.\vspace{3pt}

\noindent \textbf{Pose visualization.}~Fig.~\ref{fig:keypoint_visualize} qualitatively compares the predicted and reference 2D joint locations on MM-Fi at several compression ratios. The visualization complements, but does not replace, the PCK and MPJPE results in Tables~\ref{tab:compressionTask} and~\ref{tab:HPEtasks}. As $\eta$ increases, the main body configuration remains recognizable in the shown examples, while fine joint locations become less accurate.

\subsection{Ablation Studies}\label{sec:eval_ablation}

\begin{table}[ht!]
\centering
\small
\caption{PCK$_{20}$ (\%) versus $K$ and $\lambda$ on MM-Fi (Best in \textbf{bold}; second best underlined)}
\begin{tabular}{c|cccc}
\hline
\textbf{Weight $\lambda$} & \textbf{$K = 64$} & \textbf{$K = 128$} & \textbf{$K = 256$} & \textbf{$K = 512$}  \\
\hline
0.10 & 36.8 & 41.8 & \textbf{46.5} & 44.1  \\
0.25 & 40.5 & 42.1 & 44.3 & 43.6  \\
0.50 & 41.8 & \underline{45.7} & 41.5 & 39.3  \\
0.75 & 39.7 & 42.6 & 39.6 & 42.6  \\
1.00 & 37.5 & 42.5 & 41.5 & 43.3  \\
\hline
\end{tabular}
\label{tab:sensitivity}
\end{table}

\noindent\textbf{Hyperparameter sensitivity.} Table~\ref{tab:sensitivity} varies $K$ and adversarial weight $\lambda$ at the compression ratio $\eta=86$. The highest PCK$_{20}$ is $46.5\%$ at ($K=256$, $\lambda=0.1$), and the second highest is $45.7\%$ at ($K=128$, $\lambda=0.5$). The lowest reported value is $36.8\%$ at $K=64$, $\lambda=0.1$.

\begin{table}[ht]
    \centering
    \caption{\textcolor{black}{Ablation Study of Loss Function Components}}
    \small
    \begin{tabular}{cccc}
\hline 
 \textbf{Configuration} & \textbf{NMSE $\downarrow$} & 
$\textbf{PCK}_{20} \uparrow$ & $\textbf{MPJPE} \downarrow$ \tabularnewline
\hline 
$\mathtt{TinySense+}$ & -12.43 & 44.01 & 170.17  \tabularnewline
$\mathtt{TinySense}$ & -13.74 & 41.21 & 170.98  \tabularnewline
w/o $\mathcal{L}_{\mathtt{GAN}}$ & -13.12 & 38.50 & 179.45  \tabularnewline
w/o $\mathcal{L}_{\mathtt{keypoint}}$ & -14.50 & 29.15 & 205.10  \tabularnewline
\hline 
    \end{tabular}
    \label{tab:LossFunc}
\end{table}

\noindent\textbf{Loss components.} Table~\ref{tab:LossFunc} compares the base model and its loss ablations at $\eta=215$, with the full $\mathtt{TinySense+}$ result included for reference. As can be seen, removing the adversarial term decreases PCK$_{20}$ from $41.21\%$ to $38.50\%$. Removing keypoint supervision improves NMSE from $-13.74$ to $-14.50$~dB but reduces PCK$_{20}$ to $29.15\%$. Thus, signal reconstruction alone does not preserve the information required for accurate HPE. The $\mathtt{TinySense+}$ row uses its own coded rate, $\eta=535$, as detailed in Table~\ref{tab:tinysense_plus_component_ablation}.\vspace{3pt}

\noindent\textbf{Components of $\mathtt{TinySense+}$.} Table~\ref{tab:tinysense_plus_component_ablation} separates coding, rate-aware training, and prefix completion. Entropy coding reduces cost from $8.00$ to $4.82$ bits/index and increases $\eta$ from $215$ to $356$, while preserving NMSE, PCK$_{20}$, and MPJPE exactly. The entropy-only and no-rate-loss rows report the same operating point, consistent with lossless coding of an unchanged index sequence.

On the other hand, adding $\mathcal{L}_{\mathtt{rate}}$ reduces cost to $3.21$ bits/index ($\eta=535$). PCK$_{20}$ increases from $41.21\%$ to $44.01\%$, and MPJPE decreases from $170.98$ to $170.17$, while NMSE changes from $-13.74$ to $-12.43$~dB. The resulting representation is more compressible and gives higher PCK despite lower signal-reconstruction fidelity. Relative to fixed-length coding, the bits/index decrease is $59.9\%$. PCLC then changes the transmitted content without retraining. Prefix ratios $\rho=0.75$ and $0.50$ reduce the effective cost to $2.42$ and $1.69$ bits/index, yielding $\eta=710$ and $1015$. PCK$_{20}$ decreases to $42.73\%$ and $40.36\%$, respectively. The half-prefix setting therefore retains useful pose information while transmitting less data than the full entropy-coded sequence.\vspace{3pt}

\begin{table}[t]
    \centering
    \caption{Effect of the Rate Regularization Weight $\gamma$
    on TinySense+}
    \small
    \begin{tabular}{ccccc}
    \hline
    $\boldsymbol{\gamma}$ &
    \textbf{Bits/Index $\downarrow$} &
    $\boldsymbol{\eta}\uparrow$ &
    $\textbf{PCK}_{20}\uparrow$ &
    \textbf{NMSE $\downarrow$} \\
    \hline
    0      & 4.82 & 356  & 41.21 & -13.74 \\
    0.01   & 4.15 & 413  & 42.56 & -13.31 \\
    0.05   & 3.62 & 474  & 43.72 & -12.91 \\
    0.10   & 3.21 & 535  & 44.01 & -12.43 \\
    0.25   & 2.63 & 653  & 41.82 & -11.36 \\
    0.50   & 2.11 & 814  & 37.95 & -10.17 \\
    \hline
    \end{tabular}
    \label{tab:gamma_sensitivity}
\end{table}

\noindent\textbf{Rate-weight sensitivity.} Table~\ref{tab:gamma_sensitivity} shows that increasing $\gamma$ lowers coding cost. PCK$_{20}$ peaks at $44.01\%$ for $\gamma=0.1$, compared with $41.21\%$ without rate regularization. Stronger weights of $0.25$ and $0.50$ reduce cost to $2.63$ and $2.11$ bits/index but lower PCK$_{20}$ to $41.82\%$ and $37.95\%$. These results suggest $\gamma=0.1$ for the considered configuration, but they do not imply that one weight is optimal across all deployment budgets.\vspace{3pt}

\begin{table*}[t]
    \centering
    \caption{Ablation Study of the Training and Transmission Components
    of TinySense+ on MM-Fi}
    \small
    \setlength{\tabcolsep}{5pt}
    \begin{tabular}{lccccc}
    \hline
    \textbf{Configuration} &
    \textbf{Bits/Index $\downarrow$} &
    $\boldsymbol{\eta}\uparrow$ &
    \textbf{NMSE $\downarrow$} &
    $\textbf{PCK}_{20}\uparrow$ &
    \textbf{MPJPE $\downarrow$} \\
    \hline

    TinySense
        & 8.00 & 215
        & -13.74 & 41.21 & 170.98 \\

    TinySense + entropy model
        & 4.82 & 356
        & -13.74 & 41.21 & 170.98 \\

    TinySense+ w/o $\mathcal{L}_{\mathrm{rate}}$
        & 4.82 & 356
        & -13.74 & 41.21 & 170.98 \\

    TinySense+ w/ $\mathcal{L}_{\mathrm{rate}}$
        & 3.21 & 535
        & -12.43 & 44.01 & 170.17 \\

    TinySense+ + PCLC ($\rho=0.75$)
        & 2.42 & 710
        & -11.86 & 42.73 & 173.91 \\

    TinySense+ + PCLC ($\rho=0.50$)
        & 1.69 & 1015
        & -10.92 & 40.36 & 178.42 \\
    \hline
    \end{tabular}
    \label{tab:tinysense_plus_component_ablation}
\end{table*}

\begin{table}[ht!]
    \centering
    \small
        \caption{Analysis of the Impact of the Transformer's Sliding Window on $\mathtt{TinySense}$ Performance}
    \begin{tabular}{ccccc}
\hline 
$\text{P}\times \text{P}$  & $\textbf{NMSE\ensuremath{\downarrow}}$ & $\textbf{PCK}_{30}\ensuremath{\uparrow}$ & $\textbf{PCK}_{20}\ensuremath{\uparrow}$ & $\textbf{MPJPE\ensuremath{\downarrow}}$\tabularnewline
\hline 
$2\times2$  & -13.782 & 60.152 & 39.086 & 170.45\tabularnewline
$3\times3$   & \textbf{-14.422} & \textbf{63.158} & \textbf{46.516} & \textbf{166.87}\tabularnewline
$4\times4$   & -13.582 & 59.238 & 40.562 & 172.69\tabularnewline
$5\times5$  & \underline{-14.082} & \underline{62.881} & \underline{44.163} & \underline{168.14}\tabularnewline
\hline 
\end{tabular}
    \label{tab:slidingwindow}
\end{table}

\noindent\textbf{Recovery-window size.} Table~\ref{tab:slidingwindow} compares $2\times2$, $3\times3$, $4\times4$, and $5\times5$ windows at $\eta=86$. The $3\times3$ setting gives the best NMSE, PCK, and MPJPE. Larger windows do not improve these metrics consistently: MPJPE increases from $166.87$ to $172.69$ for $4\times4$, then decreases to $168.14$ for $5\times5$. The results support the $3\times3$ choice for this configuration. The computational efficiency is evaluated separately on the testbed.

\subsection{Rate--Accuracy Trade-offs}\label{sec:eval_rate_accuracy}

\begin{figure}[ht!] %
    \centering    \includegraphics[width=1\linewidth]{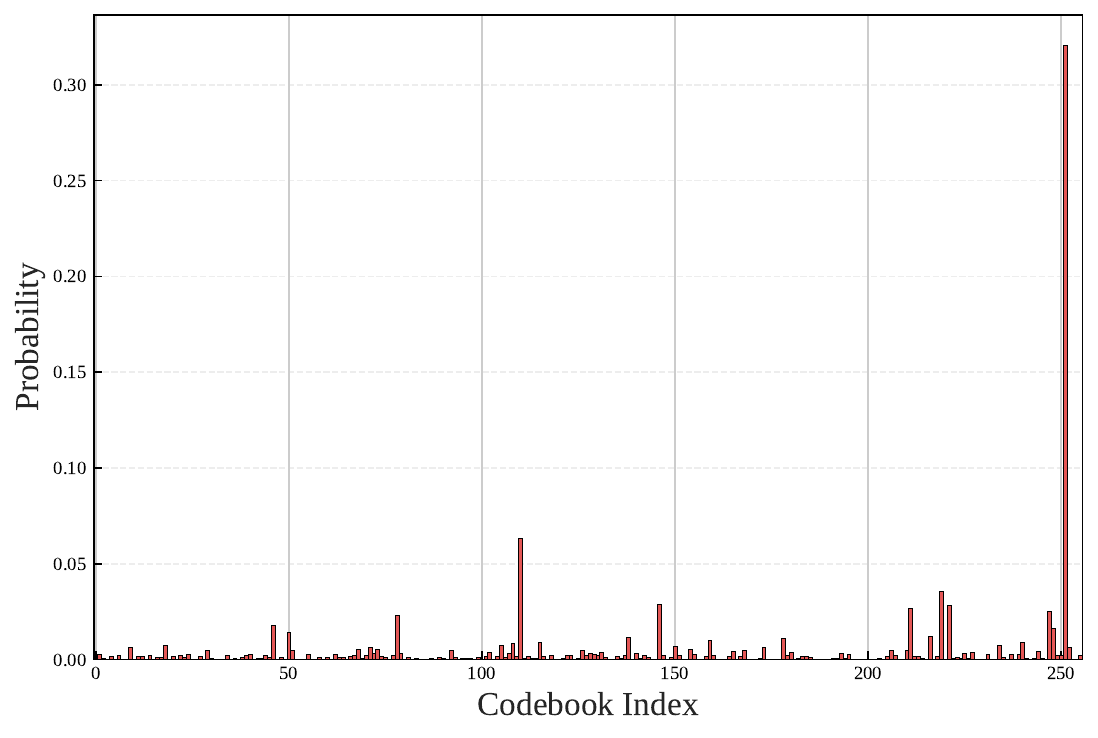} 
    
    \caption{Codebook utilization histogram of $\mathtt{TinySense+}$ ($K=256$).}
    \label{fig:codebook_util}
\end{figure}

\noindent\textbf{Codebook utilization.} To examine why $\mathtt{TinySense+}$ produces shorter bitstreams, we analyze its latent-index distribution. The base model changes the nominal rate by selecting among K-means-reduced codebooks. $\mathtt{TinySense+}$ instead regularizes the expected code length and records actual arithmetic-coded lengths. The histogram complements the rate and accuracy measurements in the component ablation.

Fig.~\ref{fig:codebook_util} shows a non-uniform codebook utilization pattern after rate-aware training. The most frequent index accounts for more than $30\%$ of the observed assignments. The histogram shows marginal index frequencies, whereas arithmetic coding uses context-dependent conditional probabilities. The histogram alone therefore does not establish the coded length or isolate the effects of the rate weight $\gamma$ and soft-assignment temperature $\tau_{\mathrm{q}}$.

\begin{figure}[ht!]
    \centering   \includegraphics[width=\linewidth]{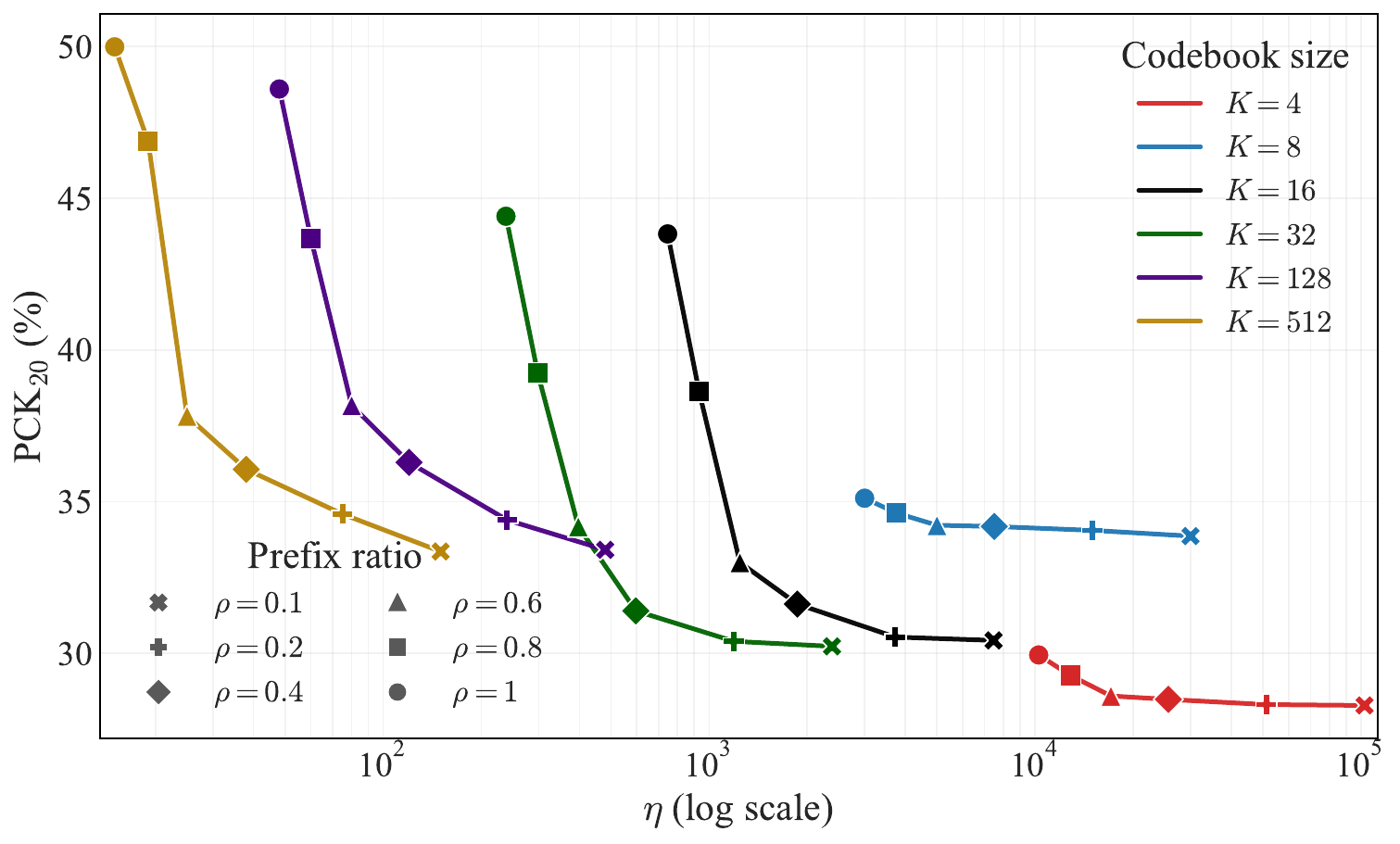}
    
    \caption{Rate--accuracy operating points of $\mathtt{TinySense+}$ under a controlled prefix-ratio sweep (Lines connect evaluated settings and $\eta$ is shown on a logarithmic axis).}
    \label{fig:rd_curve_prefix}
\end{figure}

\noindent\textbf{Prefix adaptation.}
In Fig.~\ref{fig:rd_curve_prefix}, we evaluate test-time rate adjustment by sweeping $\rho$ for fixed models with $K\in\{4,8,16,32,128,512\}$. The sweep represents externally supplied prefix budgets, rather than a time-varying wireless channel or packet-loss process. As seen, reducing the prefix from the full sequence to $10\%$ generally raises the compression ratio while lowering PCK$_{20}$, showing the cost of replacing transmitted information with generated indices. The discrete operating points demonstrate selectable rate--accuracy trade-offs. The connecting lines do not imply continuous achievable rates or guarantees for untested conditions.\vspace{3pt}

\subsection{Latency, Energy and Communication Cost}\label{sec:eval_cost}
We evaluate our framework on a real-world testbed (see Fig.~\ref{fig:setup}) under two deployment scenarios to measure end-to-end (E2E) latency, local energy consumption, and networking overhead. We compare the base model ($\mathtt{TinySense}$) and its entropy-coded extension ($\mathtt{TinySense+}$) against sensing-compression baselines (\ph{RSCNet}~\cite{20} and EfficientFi~\cite{19}) at equivalent pose estimation accuracy. Furthermore, we implement a lightweight encoder-decoder architecture consisting of 10 deep neural network (DNN) blocks, inspired by~\cite{me}, to evaluate three deployment strategies: \textbf{Local Computation (LC)}, where all computation executes entirely on the edge device; \textbf{Server-Side Computation (SC)}, where raw data is offloaded and processed remotely; and \textbf{Edge-Server Split (Ours)}, where the 3-block encoder runs locally and the remainder is deployed on the server. The considered testbed configuration uses codebook size $K=128$ and weight $\lambda=0.1$.

\begin{figure}[ht!]
    \centering
    \includegraphics[width=1.0\linewidth]{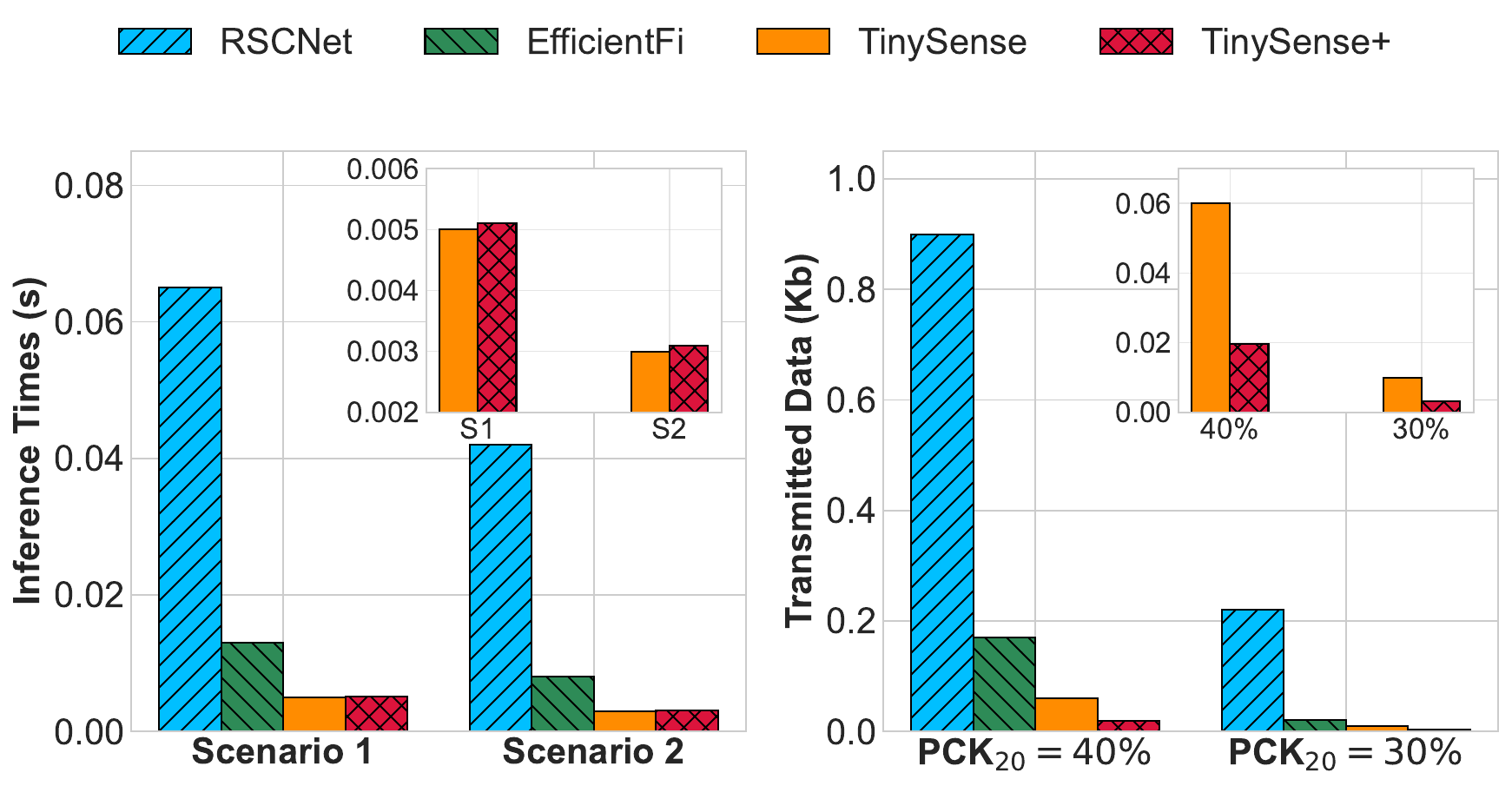} 
    
    \caption{Latency (left) and transmitted payload (right) for $\mathtt{TinySense}$, $\mathtt{TinySense+}$, and the evaluated baselines.}
    \label{fig:latency_bandwidth}
\end{figure}

\noindent \textbf{Comparison to state-of-the-art methods.}~At the matched PCK$_{20}=40\%$ point in Scenario~1, Fig.~\ref{fig:latency_bandwidth} (left) reports lower E2E latency for $\mathtt{TinySense}$ than for RSCNet~\cite{20} and EfficientFi~\cite{19}. The same ordering is observed with the Jetson Nano in Scenario~2. Fig.~\ref{fig:latency_bandwidth} (right) shows a smaller payload for $\mathtt{TinySense+}$ than for the fixed-length $\mathtt{TinySense}$ configuration at the plotted operating points. These results are specific to the evaluated devices, network, and matched-accuracy settings.\vspace{3pt}

\begin{figure}[ht!]
    \centering
    \includegraphics[width=1.0\linewidth]{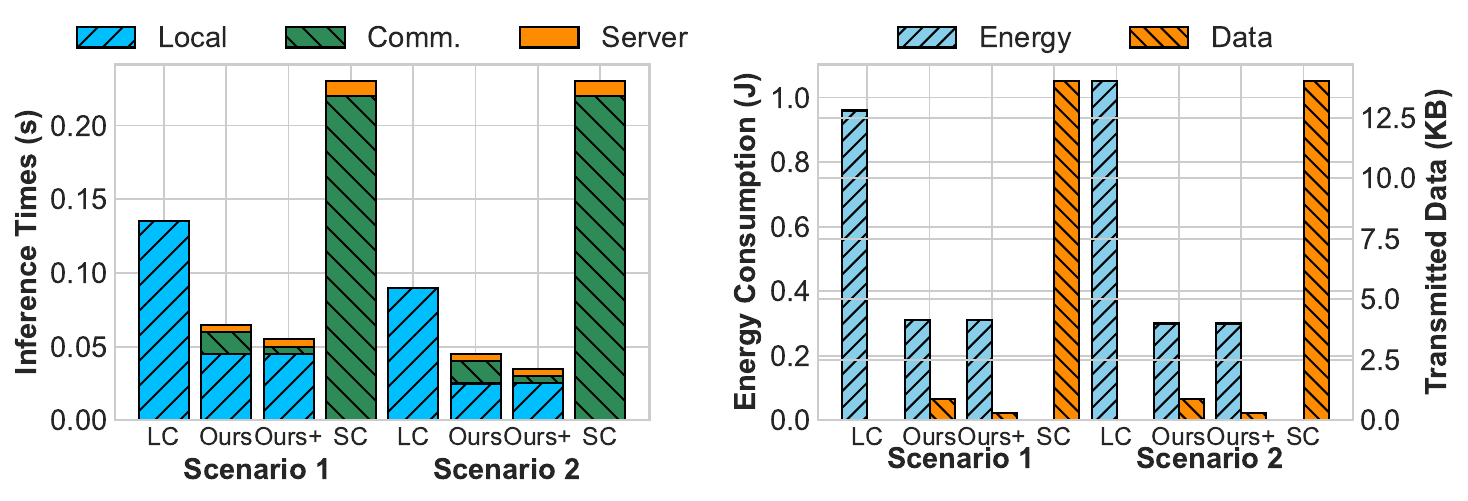} 
    
    \caption{Pipeline latency breakdown (left) and local energy versus transmitted payload (right).}
    \label{fig:inference_time}
\end{figure}

\noindent \textbf{Analysis of deployment strategies.}~The latency breakdown in Fig.~\ref{fig:inference_time} (left) separates local computation, communication, and server computation. Arithmetic coding adds computation to $\mathtt{TinySense+}$, but the reduction in communication time is larger in the evaluated setup, so its total E2E latency remains below the LC and SC alternatives. The result illustrates a computation--communication trade-off whose benefit depends on the hardware and link conditions.

Fig.~\ref{fig:inference_time} (right) compares local energy consumption and transmitted data volume. Among the evaluated split architectures, $\mathtt{TinySense+}$ achieves the lowest local energy consumption and smallest payload on both testbed devices. These results indicate that the additional computation required for entropy coding does not outweigh the energy benefit of reduced communication in the evaluated configurations. The measurements support the practical value of combining local CSI compression with server-side inference. Further evaluation is needed to assess how these benefits vary across hardware platforms, radio conditions, and numbers of concurrent users.

\section{Conclusion}
\label{sec:conclusion}
This work presented $\mathtt{TinySense+}$, an entropy-aware extension of $\mathtt{TinySense}$ for efficient edge--server Wi-Fi pose estimation. The framework combined task-aware vector quantization, causal entropy modeling, and differentiable rate regularization, and reused the entropy model to complete omitted latent indices for test-time rate adjustment. Evaluations on MM-Fi and WiPose demonstrated that compression efficiency depended on both latent predictability and the preservation of pose information. On MM-Fi, entropy coding and rate-aware training reduced the coding cost from $8.00$ to $3.21$ bits per index while improving PCK$_{20}$ from $41.21\%$ to $44.01\%$. Prefix transmission provided additional control over the communication--accuracy trade-off without encoder retraining. Raspberry Pi and Jetson Nano experiments assessed the associated latency, energy, and communication costs. Together, these results demonstrated the value of jointly optimizing representation, coding, and transmission length for CSI-based human pose estimation.

\bibliographystyle{IEEEtran} 
\balance
\bibliography{reference}

\end{document}